\documentclass[11pt]{article}

\usepackage[final]{acl}

\usepackage{times}
\usepackage{latexsym}

\usepackage[T1]{fontenc}

\usepackage[utf8]{inputenc}

\usepackage{microtype}

\usepackage{inconsolata}

\usepackage{graphicx}

\usepackage{booktabs} 
\usepackage{rotating}
\usepackage{footnotehyper}
\usepackage{hyperref}
\usepackage{savefnmark}
\usepackage{tablefootnote}

\usepackage{algorithm}
\usepackage{algorithmic}
\usepackage{amsmath}
\usepackage{amssymb}
\usepackage{mathtools}
\usepackage{amsthm}
\usepackage{subcaption}
\usepackage{utfsym}
\usepackage{tcolorbox}
\usepackage{pifont}

\usepackage{array}
\usepackage{listings}
\usepackage{makecell} 

\tcbuselibrary{breakable}  
\tcbuselibrary{skins}
\definecolor{PineGreen}{HTML}{007C4A}
\definecolor{OrangeRed}{HTML}{FF4500}
\definecolor{mycolor}{gray}{0.92}

\usepackage[capitalize,noabbrev]{cleveref}

\theoremstyle{plain}

\theoremstyle{definition}

\theoremstyle{remark}

\usepackage{dcolumn, array, booktabs}
\usepackage{multirow}
\usepackage{cuted}
\usepackage[table]{xcolor}
\usepackage[dvipsnames]{xcolor}
\definecolor{skyblue}{RGB}{210,235,255}
\definecolor{lightcoral}{RGB}{255,228,225}
\definecolor{lemonchiffon}{RGB}{255,250,205}

\definecolor{prompt}{RGB}{173, 216, 230}
\definecolor{prompt-frame}{RGB}{0, 51, 102}
\definecolor{prompt2}{RGB}{173, 216, 230}
\definecolor{prompt2-frame}{RGB}{0, 51, 102}
\definecolor{prompt3}{RGB}{173, 216, 230}
\definecolor{prompt3-frame}{RGB}{0, 51, 102}
\definecolor{prompt4}{RGB}{173, 216, 230}
\definecolor{prompt4-frame}{RGB}{0, 51, 102}
\definecolor{prompt5}{RGB}{173, 216, 230}
\definecolor{prompt5-frame}{RGB}{0, 51, 102}
\usepackage{fontawesome5}
\tcbset{
    prompt_func/.style={
        enhanced,
        breakable,
        colback=prompt,        
        colframe=prompt-frame,            
        fontupper=\normalsize,               
        boxrule=1pt,                       
        arc=4mm,                           
        left=1mm, right=1mm, top=1mm, bottom=1mm, 
        boxsep=4pt,                        
        before skip=10pt, after skip=10pt, 
        overlay={
            \node[anchor=north west, xshift=4pt, text=white] at (frame.north west) {\faDatabase};
        },
        title={~~~~~~~\textbf{Prompt used with GPT-4o to generate annotation for Element Grounding with functional setting}},
        coltitle=white,
        fonttitle=\bfseries\small
    }
}

\tcbset{
    full_content_generation/.style={
        enhanced,
        breakable,
        colback=prompt,        
        colframe=prompt-frame,            
        fontupper=\normalsize,               
        boxrule=1pt,                       
        arc=4mm,                           
        left=1mm, right=1mm, top=1mm, bottom=1mm, 
        boxsep=4pt,                        
        before skip=10pt, after skip=10pt, 
        title={\faDatabase~~~\textbf{Prompt: Full Content Generation}},
        coltitle=white,
        fonttitle=\bfseries\small
    }
}
\tcbset{
    filter_figures/.style={
        enhanced,
        breakable,
        colback=prompt,        
        colframe=prompt-frame,            
        fontupper=\normalsize,               
        boxrule=1pt,                       
        arc=4mm,                           
        left=1mm, right=1mm, top=1mm, bottom=1mm, 
        boxsep=4pt,                        
        before skip=10pt, after skip=10pt, 
        title={\faDatabase~~~\textbf{Prompt: Filter Figures}},
        coltitle=white,
        fonttitle=\bfseries\small
    }
}
\tcbset{
    section_generation/.style={
        enhanced,
        breakable,
        colback=prompt,        
        colframe=prompt-frame,            
        fontupper=\normalsize,               
        boxrule=1pt,                       
        arc=4mm,                           
        left=1mm, right=1mm, top=1mm, bottom=1mm, 
        boxsep=4pt,                        
        before skip=10pt, after skip=10pt, 
        title={\faDatabase~~~\textbf{Prompt: Section Generation}},
        coltitle=white,
        fonttitle=\bfseries\small
    }
}
\tcbset{
    html_generation/.style={
        enhanced,
        breakable,
        colback=prompt,        
        colframe=prompt-frame,            
        fontupper=\normalsize,               
        boxrule=1pt,                       
        arc=4mm,                           
        left=1mm, right=1mm, top=1mm, bottom=1mm, 
        boxsep=4pt,                        
        before skip=10pt, after skip=10pt, 
        title={\faDatabase~~~\textbf{Prompt: HTML Generation}},
        coltitle=white,
        fonttitle=\bfseries\small
    }
}
\tcbset{
    full_content_review/.style={
        enhanced,
        breakable,
        colback=prompt,        
        colframe=prompt-frame,            
        fontupper=\normalsize,               
        boxrule=1pt,                       
        arc=4mm,                           
        left=1mm, right=1mm, top=1mm, bottom=1mm, 
        boxsep=4pt,                        
        before skip=10pt, after skip=10pt, 
        title={\faDatabase~~~\textbf{Prompt: Full Content Review}},
        coltitle=white,
        fonttitle=\bfseries\small
    }
}
\tcbset{
    full_content_revise/.style={
        enhanced,
        breakable,
        colback=prompt,        
        colframe=prompt-frame,            
        fontupper=\normalsize,               
        boxrule=1pt,                       
        arc=4mm,                           
        left=1mm, right=1mm, top=1mm, bottom=1mm, 
        boxsep=4pt,                        
        before skip=10pt, after skip=10pt, 
        title={\faDatabase~~~\textbf{Prompt: Full Content Revise}},
        coltitle=white,
        fonttitle=\bfseries\small
    }
}
\tcbset{
    text_content_generation/.style={
        enhanced,
        breakable,
        colback=prompt,        
        colframe=prompt-frame,            
        fontupper=\normalsize,               
        boxrule=1pt,                       
        arc=4mm,                           
        left=1mm, right=1mm, top=1mm, bottom=1mm, 
        boxsep=4pt,                        
        before skip=10pt, after skip=10pt, 
        title={\faDatabase~~~\textbf{Prompt: Text Content Generation}},
        coltitle=white,
        fonttitle=\bfseries\small
    }
}
\tcbset{
    html_review/.style={
        enhanced,
        breakable,
        colback=prompt,        
        colframe=prompt-frame,            
        fontupper=\normalsize,               
        boxrule=1pt,                       
        arc=4mm,                           
        left=1mm, right=1mm, top=1mm, bottom=1mm, 
        boxsep=4pt,                        
        before skip=10pt, after skip=10pt, 
        title={\faDatabase~~~\textbf{Prompt: HTML Review}},
        coltitle=white,
        fonttitle=\bfseries\small
    }
}
\tcbset{
    html_revise/.style={
        enhanced,
        breakable,
        colback=prompt,        
        colframe=prompt-frame,            
        fontupper=\normalsize,               
        boxrule=1pt,                       
        arc=4mm,                           
        left=1mm, right=1mm, top=1mm, bottom=1mm, 
        boxsep=4pt,                        
        before skip=10pt, after skip=10pt, 
        title={\faDatabase~~~\textbf{Prompt: HTML Revise}},
        coltitle=white,
        fonttitle=\bfseries\small
    }
}
\tcbset{
    prompt_aethetic_element/.style={
        enhanced,
        breakable,
        colback=prompt,        
        colframe=prompt-frame,            
        fontupper=\normalsize,               
        boxrule=1pt,                       
        arc=4mm,                           
        left=1mm, right=1mm, top=1mm, bottom=1mm, 
        boxsep=4pt,                        
        before skip=10pt, after skip=10pt, 
        title={\faDatabase~~~\textbf{Prompt: Element Quality Judge}},
        coltitle=white,
        fonttitle=\bfseries\small
    }
}

\tcbset{
    prompt_aethetic_Layout/.style={
        enhanced,
        breakable,
        colback=prompt,        
        colframe=prompt-frame,            
        fontupper=\normalsize,               
        boxrule=1pt,                       
        arc=4mm,                           
        left=1mm, right=1mm, top=1mm, bottom=1mm, 
        boxsep=4pt,                        
        before skip=10pt, after skip=10pt, 
        title={\faDatabase~~~\textbf{Prompt: Layout Balance Judge}},
        coltitle=white,
        fonttitle=\bfseries\small
    }
}

\tcbset{prompt_aethetic_engagement/.style={
        enhanced,
        breakable,
        colback=prompt,        
        colframe=prompt-frame,            
        fontupper=\normalsize,               
        boxrule=1pt,                       
        arc=4mm,                           
        left=1mm, right=1mm, top=1mm, bottom=1mm, 
        boxsep=4pt,                        
        before skip=10pt, after skip=10pt, 
        title={\faDatabase~~~\textbf{Prompt: Engagement Judge}},
        coltitle=white,
        fonttitle=\bfseries\small
    }
}

\tcbset{prompt_info_clarity/.style={
        enhanced,
        breakable,
        colback=prompt,        
        colframe=prompt-frame,            
        fontupper=\normalsize,               
        boxrule=1pt,                       
        arc=4mm,                           
        left=1mm, right=1mm, top=1mm, bottom=1mm, 
        boxsep=4pt,                        
        before skip=10pt, after skip=10pt, 
        title={\faDatabase~~~\textbf{Prompt: Clarity Judge}},
        coltitle=white,
        fonttitle=\bfseries\small
    }
}

\tcbset{prompt_info_content/.style={
        enhanced,
        breakable,
        colback=prompt,        
        colframe=prompt-frame,            
        fontupper=\normalsize,               
        boxrule=1pt,                       
        arc=4mm,                           
        left=1mm, right=1mm, top=1mm, bottom=1mm, 
        boxsep=4pt,                        
        before skip=10pt, after skip=10pt, 
        title={\faDatabase~~~\textbf{Prompt: Content Completeness Judge}},
        coltitle=white,
        fonttitle=\bfseries\small
    }
}

\tcbset{prompt_info_logic/.style={
        enhanced,
        breakable,
        colback=prompt,        
        colframe=prompt-frame,            
        fontupper=\normalsize,               
        boxrule=1pt,                       
        arc=4mm,                           
        left=1mm, right=1mm, top=1mm, bottom=1mm, 
        boxsep=4pt,                        
        before skip=10pt, after skip=10pt, 
        title={\faDatabase~~~\textbf{Prompt: Logical Flow Judge}},
        coltitle=white,
        fonttitle=\bfseries\small
    }
}

\tcbset{prompt_verbatim_question/.style={
        enhanced,
        breakable,
        colback=prompt,        
        colframe=prompt-frame,            
        fontupper=\normalsize,               
        boxrule=1pt,                       
        arc=4mm,                           
        left=1mm, right=1mm, top=1mm, bottom=1mm, 
        boxsep=4pt,                        
        before skip=10pt, after skip=10pt, 
        title={\faDatabase~~~\textbf{Prompt: Generate Verbatim QA}},
        coltitle=white,
        fonttitle=\bfseries\small
    }
}

\tcbset{prompt_interpretive_question/.style={
        enhanced,
        breakable,
        colback=prompt,        
        colframe=prompt-frame,            
        fontupper=\normalsize,               
        boxrule=1pt,                       
        arc=4mm,                           
        left=1mm, right=1mm, top=1mm, bottom=1mm, 
        boxsep=4pt,                        
        before skip=10pt, after skip=10pt, 
        title={\faDatabase~~~\textbf{Prompt: Generate Interpretive QA}},
        coltitle=white,
        fonttitle=\bfseries\small
    }
}

\tcbset{prompt_answer_agent/.style={
        enhanced,
        breakable,
        colback=prompt,        
        colframe=prompt-frame,            
        fontupper=\normalsize,               
        boxrule=1pt,                       
        arc=4mm,                           
        left=1mm, right=1mm, top=1mm, bottom=1mm, 
        boxsep=4pt,                        
        before skip=10pt, after skip=10pt, 
        title={\faDatabase~~~\textbf{Prompt: Answer Questions}},
        coltitle=white,
        fonttitle=\bfseries\small
    }
}

\tcbset{prompt_4o_image/.style={
        enhanced,
        breakable,
        colback=prompt,        
        colframe=prompt-frame,            
        fontupper=\normalsize,               
        boxrule=1pt,                       
        arc=4mm,                           
        left=1mm, right=1mm, top=1mm, bottom=1mm, 
        boxsep=4pt,                        
        before skip=10pt, after skip=10pt, 
        title={\faDatabase~~~\textbf{Prompt: \texttt{4o-Image}}},
        coltitle=white,
        fonttitle=\bfseries\small
    }
}

\tcbset{prompt_owl_4o/.style={
        enhanced,
        breakable,
        colback=prompt,        
        colframe=prompt-frame,            
        fontupper=\normalsize,               
        boxrule=1pt,                       
        arc=4mm,                           
        left=1mm, right=1mm, top=1mm, bottom=1mm, 
        boxsep=4pt,                        
        before skip=10pt, after skip=10pt, 
        title={\faDatabase~~~\textbf{Prompt: \texttt{OWL-4o}}},
        coltitle=white,
        fonttitle=\bfseries\small
    }
}

\tcbset{prompt_4o_html/.style={
        enhanced,
        breakable,
        colback=prompt,        
        colframe=prompt-frame,            
        fontupper=\normalsize,               
        boxrule=1pt,                       
        arc=4mm,                           
        left=1mm, right=1mm, top=1mm, bottom=1mm, 
        boxsep=4pt,                        
        before skip=10pt, after skip=10pt, 
        title={\faDatabase~~~\textbf{Prompt: \texttt{4o-HTML}}},
        coltitle=white,
        fonttitle=\bfseries\small
    }
}

\tcbset{prompt_parser_summarizer/.style={
        enhanced,
        breakable,
        colback=prompt,        
        colframe=prompt-frame,            
        fontupper=\normalsize,               
        boxrule=1pt,                       
        arc=4mm,                           
        left=1mm, right=1mm, top=1mm, bottom=1mm, 
        boxsep=4pt,                        
        before skip=10pt, after skip=10pt, 
        title={\faDatabase~~~\textbf{Prompt: Paper Summarizer}},
        coltitle=white,
        fonttitle=\bfseries\small
    }
}

\tcbset{prompt_parser_filter/.style={
        enhanced,
        breakable,
        colback=prompt,        
        colframe=prompt-frame,            
        fontupper=\normalsize,               
        boxrule=1pt,                       
        arc=4mm,                           
        left=1mm, right=1mm, top=1mm, bottom=1mm, 
        boxsep=4pt,                        
        before skip=10pt, after skip=10pt, 
        title={\faDatabase~~~\textbf{Prompt: Figure Filter}},
        coltitle=white,
        fonttitle=\bfseries\small
    }
}

\tcbset{prompt_planner_matching/.style={
        enhanced,
        breakable,
        colback=prompt,        
        colframe=prompt-frame,            
        fontupper=\normalsize,               
        boxrule=1pt,                       
        arc=4mm,                           
        left=1mm, right=1mm, top=1mm, bottom=1mm, 
        boxsep=4pt,                        
        before skip=10pt, after skip=10pt, 
        title={\faDatabase~~~\textbf{Prompt: Asset Matching}},
        coltitle=white,
        fonttitle=\bfseries\small
    }
}

\tcbset{prompt_planner_painter/.style={
        enhanced,
        breakable,
        colback=prompt,        
        colframe=prompt-frame,            
        fontupper=\normalsize,               
        boxrule=1pt,                       
        arc=4mm,                           
        left=1mm, right=1mm, top=1mm, bottom=1mm, 
        boxsep=4pt,                        
        before skip=10pt, after skip=10pt, 
        title={\faDatabase~~~\textbf{Prompt: Painter}},
        coltitle=white,
        fonttitle=\bfseries\small
    }
}

\tcbset{prompt_planner_commenter/.style={
        enhanced,
        breakable,
        colback=prompt,        
        colframe=prompt-frame,            
        fontupper=\normalsize,               
        boxrule=1pt,                       
        arc=4mm,                           
        left=1mm, right=1mm, top=1mm, bottom=1mm, 
        boxsep=4pt,                        
        before skip=10pt, after skip=10pt, 
        title={\faDatabase~~~\textbf{Prompt: Commenter}},
        coltitle=white,
        fonttitle=\bfseries\small
    }
}

\tcbset{
    vlm_aesthetics_judge/.style={
        enhanced,
        breakable,
        colback=prompt,        
        colframe=prompt-frame,            
        fontupper=\normalsize,               
        boxrule=1pt,                       
        arc=4mm,                           
        left=1mm, right=1mm, top=1mm, bottom=1mm, 
        boxsep=4pt,                        
        before skip=10pt, after skip=10pt, 
        title={\faDatabase~~~\textbf{Prompt: Aesthetics Quality Judge}},
        coltitle=white,
        fonttitle=\bfseries\small
    }
}
\tcbset{
    vlm_element_judge/.style={
        enhanced,
        breakable,
        colback=prompt,        
        colframe=prompt-frame,            
        fontupper=\normalsize,               
        boxrule=1pt,                       
        arc=4mm,                           
        left=1mm, right=1mm, top=1mm, bottom=1mm, 
        boxsep=4pt,                        
        before skip=10pt, after skip=10pt, 
        title={\faDatabase~~~\textbf{Prompt: Element Quality Judge}},
        coltitle=white,
        fonttitle=\bfseries\small
    }
}

\tcbset{
    vlm_layout_judge/.style={
        enhanced,
        breakable,
        colback=prompt,        
        colframe=prompt-frame,            
        fontupper=\normalsize,               
        boxrule=1pt,                       
        arc=4mm,                           
        left=1mm, right=1mm, top=1mm, bottom=1mm, 
        boxsep=4pt,                        
        before skip=10pt, after skip=10pt, 
        title={\faDatabase~~~\textbf{Prompt: Layout Quality Judge}},
        coltitle=white,
        fonttitle=\bfseries\small
    }
}

\newtcolorbox{definitionbox}{
  colback=gray!10,
  colframe=gray!80,
  boxrule=1pt,
  width=\textwidth,
  enlarge left by=0mm,
  enlarge right by=0mm,
  float*=t,  
  boxsep=1pt,
  arc=2.5mm,
  halign=left,
}

\newtcolorbox{notebox}[1]{
  breakable,
  colback=gray!5,
  colframe=gray!60,
  boxrule=0.8pt,
  arc=2mm,
  boxsep=2pt,
  left=6pt, right=6pt, top=4pt, bottom=4pt,
  title={\textbf{NOTE: #1}},
  fonttitle=\small,
  fontupper=\footnotesize,
  before upper={\setlength{\itemsep}{0pt}\setlength{\parsep}{0pt}\setlength{\topsep}{2pt}},
  halign=left,
}

\title{FundingClaw: Automated Grant Proposal Writing Agents System}

\author{Ruizhi Liu \and Siru Ge \and Siran Wang \and Qianli Ma \and Zhipeng Zhang\thanks{Corresponding author: zhipengzhang@sjtu.edu.cn} \\
  Shanghai Jiao Tong University\\
}

\begin{document}
\maketitle

\begin{abstract}
Grant proposal writing remains a manual, labor-intensive bottleneck: proposals must satisfy rigid, agency-specific templates, and the same intellectual content must be repeatedly re-encoded into incompatible formats across funding organizations. We introduce \textbf{FundingClaw}, a multi-agent framework built on template-driven skill selection, where a Template Parser ingests any agency's specification and selects the appropriate writing skills for each stage, so that one research brief can produce compliant proposals for NSFC, NSF, KAKENHI, or ERC without modifying generation logic. A blueprint-skeleton strategy drafts a structured outline before any prose is written, mitigating the semantic drift that plagues long-form generation. Three sequential stages produce the rationale, research content, and research foundation sections, with multi-source verification of the PI's record and optional PDF-grounded extraction of representative works. Optional human feedback checkpoints and a closed-loop self-review complete the pipeline. Evaluated on ten research topics with an eight-dimension rubric, FundingClaw outperforms the baseline, narrowing the gap with human level for laborious funding proposal. The code will be released upon acceptance.
\end{abstract}

\section{Introduction}

\indent Grant proposals share a common intellectual skeleton with research papers, both requiring a research vision, a field survey, and grounding in evidence. LLM-based agents have shown these sub-tasks can be partially automated, with agentic systems applied to literature synthesis~\cite{wang2024autosurvey,li2025search}, experiment design~\cite{huang2024understanding}, and manuscript drafting~\cite{wan2025cognitive,weng2025cycleresearcher}, backed by multi-agent collaboration frameworks~\cite{wu2024autogen,tran2025multi,yao2023react,hong2024metagpt}. The emerging Automated Research paradigm envisions agents handling the full workflow from hypothesis generation to camera-ready paper~\cite{yang2026aris,li2025webthinker,lu2024aiscientist}, compressing the research cycle and repositioning the researcher as strategic supervisor~\cite{wang2024survey,xi2025rise,zhao2023survey}.

Grant proposals, however, differ fundamentally from research papers in two respects that existing agentic systems have not addressed. First, proposals place far greater weight on implementation feasibility and long-term planning: reviewers evaluate not only the scientific vision but also the concrete task decomposition, risk mitigation, resource allocation, and multi-year milestones, requiring a depth of structured argumentation that single-pass generation consistently fails to produce at the required quality~\cite{wan2025cognitive,weng2025cycleresearcher}. Second, proposals are tightly constrained by agency-specific templates that vary substantially across countries and funding bodies: NSFC, NSF, ERC, and KAKENHI each impose distinct section hierarchies, per-section length budgets, citation scopes, and compliance rules. Direct end-to-end generation cannot reliably satisfy these constraints, and existing pipeline systems such as ARIS~\cite{yang2026aris}, CycleResearcher~\cite{weng2025cycleresearcher}, and AutoSurvey~\cite{wang2024autosurvey} are designed for conventional research papers and fall short in three respects when applied to grant proposals: they provide limited support for the structured feasibility and planning content that reviewers evaluate, their literature survey components do not reach the systematic depth and breadth that grant rationale sections require, and they lack any mechanism for verifying or summarizing the PI's publication record, prior projects, and research foundation, which are mandatory and heavily scrutinized components of every grant submission. Manually redesigning the pipeline for each new agency defeats the purpose of automation and scales poorly. We call this the \textit{one-to-many template mapping problem}. Together, these gaps define a challenge that no existing automated writing system addresses.

To address these two challenges, we introduce \textbf{FundingClaw}, a multi-agent framework for automated grant proposal writing. To tackle feasibility depth and long-form coherence, a blueprint-skeleton strategy drafts a structured outline before any prose is written and expands each section under its constraint, countering semantic drift; three dedicated sequential stages then produce content of qualitatively distinct types: \textbf{Research Rationale} constructs a literature-grounded motivation via a Literature Prefetch, parallel Significance and Landscape analyzers, and writer skills; \textbf{Research Content} produces tasks, scientific questions, and measurable objectives under a Citation Scope Gate; \textbf{Research Foundation} cross-validates the PI's record against Google Scholar, DBLP, and Semantic Scholar and optionally performs PDF-grounded extraction~\cite{wang2024mineruopensourcesolutionprecise,Docling,blecher2023nougat} of representative works. To tackle template heterogeneity, a Template Parser ingests any agency's template document, extracting its section hierarchy, per-section page estimates, and formatting constraints, then drives template-driven skill selection that filters and assigns the appropriate stage-level writing skills to each proposal stage. The same research brief thereby generates compliant proposals for NSFC, NSF, KAKENHI, ERC, or any novel agency. Optional human feedback checkpoints at stage boundaries and a final self-review agent complete the pipeline.

The contributions of this paper are summarized as follows:

\begin{itemize}
    \item We introduce \textbf{FundingClaw}, the end-to-end multi-agent framework for automated grant proposal writing, covering a pipeline with embedded literature retrieval, multi-source applicant verification, and PDF-grounded prior-work exposition across heterogeneous funding agencies.
                                              
    \item We propose \textbf{template-driven skill selection}, which decouples agency regulations from generation logic. Agency rules are externalized as parsed data rather than hardcoded control flow, and writing skills target agency-independent functional roles rather than agency-specific section titles. A Template Parser ingests any agency's specification and routes each functional role to the appropriate skill, so one generator serves many agencies without modifying generation logic.

    \item We introduce a blueprint-skeleton generation strategy that drafts a structured outline before any prose is written and expands each section progressively under its constraint, mitigating semantic drift and cross-section misalignment in long-form generation.

    \item We design stage-boundary human feedback checkpoints that allow the PI to inspect generated content and issue natural-language correction instructions at each pipeline stage, enabling timely alignment with the researcher's preferences without restarting the pipeline.                     

    \item We design an eight-dimension, 78-item evaluation rubric for grant proposal quality and instantiate it across four agency templates (NSFC, NSF, ERC, KAKENHI). All four share the same dimensional structure and scoring levels, with per-agency point allocations adjusted to reflect each template's emphasis, providing a reusable and cross-agency benchmark for automated proposal evaluation.
\end{itemize}
 
\section{Related Works}

\noindent\textbf{Agentic Systems and Scientific Writing.}~
Large language models (LLMs) have been extended from standalone generators into agentic systems~\cite{wang2024survey,xi2025rise,xie2024large,zhao2023survey} capable of autonomous planning~\cite{huang2024understanding,sun2023adaplanner}, tool use~\cite{qu2025tool,shi2025tool}, reasoning~\cite{fu2023improving}, and multi-agent collaboration~\cite{tran2025multi}, formalized by frameworks such as ReAct~\cite{yao2023react}, LangChain~\cite{Chase_LangChain_2022}, and AutoGen~\cite{wu2024autogen}, with retrieval~\cite{li2025webthinker,li2025search} and self-reflection~\cite{shinn2023reflexion} further strengthening reliability. In the scientific domain, these capabilities have been applied to literature synthesis~\cite{wang2024autosurvey}, manuscript writing and review~\cite{weng2025cycleresearcher,ma2026paper2rebuttal,zhu2025deepreviewimprovingllmbasedpaper}, and experimental code reproduction~\cite{seo2025paper2code}; ARIS~\cite{yang2026aris} further orchestrates adversarial multi-agent collaboration over a library of reusable skill modules with claim verification. For the research foundation content that grant proposals require, layout-aware parsers such as MinerU~\cite{wang2024mineruopensourcesolutionprecise} and Docling~\cite{Docling} recover figures, tables, and reading order from publication PDFs, and corpora such as S2ORC~\cite{lo2020s2orc} supply structured, verifiable metadata. FundingClaw extends this line of work to grant proposal generation, a setting that imposes qualitatively different constraints, including regulatory compliance, anti-hallucination guarantees on PI credentials, and agency-specific argumentation structures that prior agentic writing systems have not addressed.

\noindent\textbf{Structured Document Templates and Workflow Adaptation.}~
Grant proposals differ from free-form scientific writing in that each funding agency enforces a rigid template specifying sections, page budgets, citation scope, and formatting conventions that vary across countries and grant types. Prior work has shown that multi-agent systems can leverage semantic template schemas to guide structured content creation for administrative documents~\cite{musumeci2024semistruct}, and AFlow~\cite{zhang2024aflow} established that dynamic workflow construction via execution feedback outperforms static manual design. Proposals span 20--40 pages of tightly regulated text, a scale at which agent workflows exhibit content drift and internal inconsistency~\cite{wan2025cognitive} that iterative self-refinement~\cite{madaan2023selfrefine} cannot fully resolve. FundingClaw extends these insights to the regulatory-template setting: a template parser deterministically maps parsed agency specifications onto writing skills, and blueprint-skeleton aligned serial generation with optional human checkpoints counters drift. To our knowledge, it is the first end-to-end system to target this setting.

Table~\ref{tab:system_comparison} in Appendix~\ref{app:system_comparison} summarises the key capability differences between FundingClaw and the most closely related automated writing systems.

\noindent\textbf{Concurrent Work.}~Synteract~\cite{synteract2026} proposes an experience-based interaction paradigm for stepwise task success via structured inter-agent advice exchange. Unlike our work, it targets general task execution rather than regulatory document generation.

\section{Method}
\label{sec:pageagent}

\begin{figure*}[t]
    \centering
    \includegraphics[width=\textwidth]{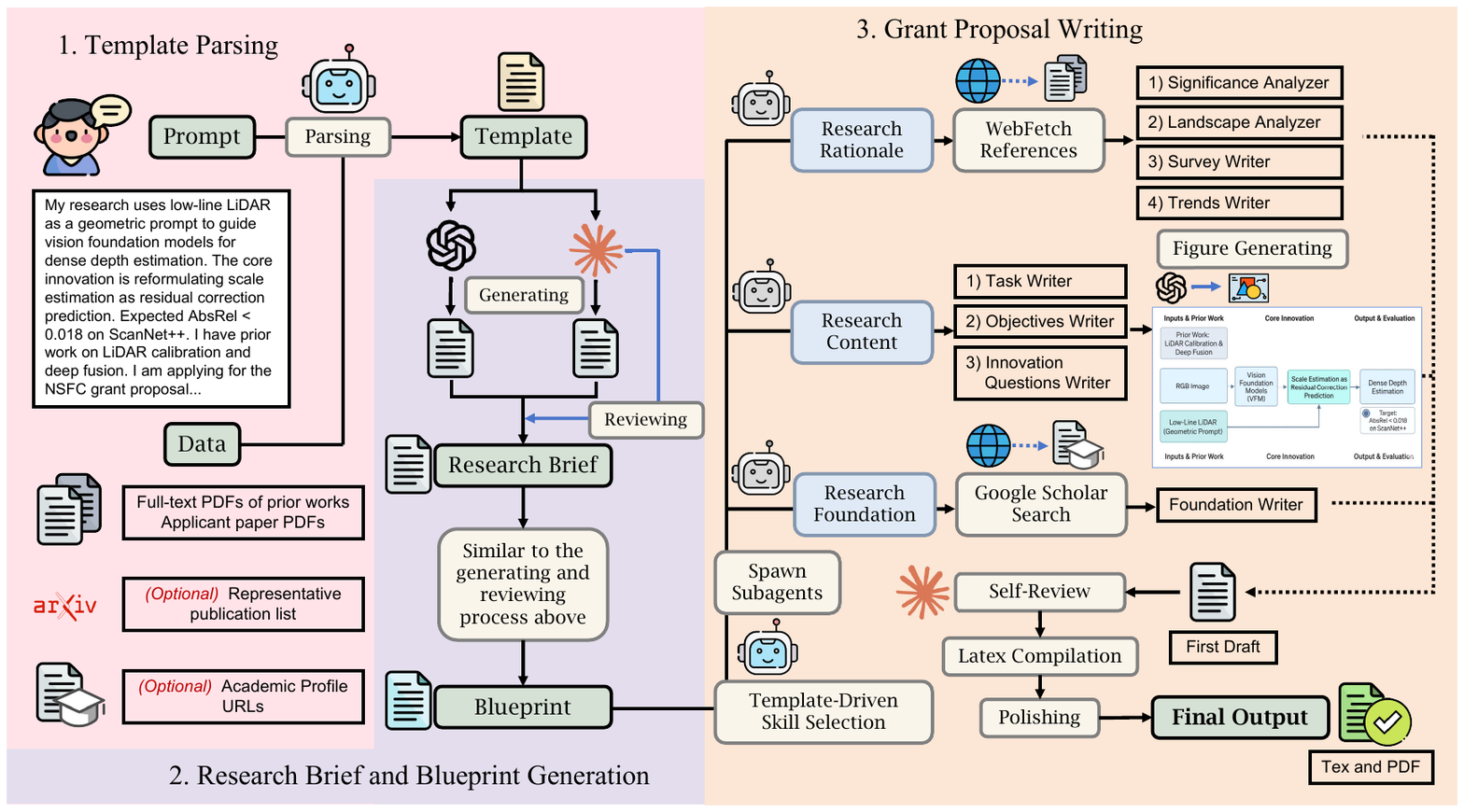} 
    \caption{\textbf{FundingClaw system workflow.} \textbf{(1) Template Parsing}: the user's prompt and agency template are parsed into a structured template specification. \textbf{(2) Research Brief \& Blueprint Generation}: a generator--reviewer pair conditioned on the template, prior-work PDFs, an optional representative publication list, and optional profile URLs produces a Research Brief and a shared Blueprint. \textbf{(3) Grant Proposal Writing}: guided by template-driven skill selection, three subagents run sequentially: \textit{Research Rationale} fetches web references and passes them through Significance Analyzer, Landscape Analyzer, Survey Writer, and Trends Writer, then generates figures; \textit{Research Content} runs Task Writer, Objectives Writer, and Innovation Questions Writer; \textit{Research Foundation} queries Google Scholar and runs Foundation Writer and Translation Writer. The combined draft undergoes self-review, LaTeX compilation, and polishing to produce the final output.}
    \label{fig:pipeline}
\end{figure*}

\textbf{FundingClaw} is an automated funding proposal writing agent with optional human checkpoints and is able to adapt to different templates. The pipeline is governed by two cross-cutting mechanisms: template-driven skill selection~(Sec.~\ref{sec:reorch}), which maps each target agency's template onto the appropriate writing skills, and blueprint-skeleton aligned serial generation~(Sec.~\ref{sec:blueprint}), which anchors all stages to a shared outline to prevent semantic drift. These mechanisms govern the following three sequential writing stages~(Sec.~\ref{sec:proposal_writing}) and a quality control loop~(Sec.~\ref{sec:quality_control}). The overview of our system is illustrated in Fig.~\ref{fig:pipeline}.

\subsection{Template-Driven Skill Selection}
\label{sec:reorch}

The core challenge in multi-agency proposal generation is that agency regulations including section hierarchies, length budgets, citation scopes, and compliance rules differ substantially across funding bodies. Conventional pipelines \textbf{hardcode} these rules into generation logic, so adapting to a new agency requires reimplementing its sections and constraints throughout the codebase. FundingClaw addresses this by \textit{decoupling agency regulation from generation logic}: regulations are externalized as parsed data, while writing skills target agency-independent functional roles (e.g., motivation survey, task decomposition, prior-work exposition) rather than agency-specific section titles. Adding a new agency requires only supplying its specification document; the generation logic remains unchanged.

\begin{figure}[h]
\centering
\includegraphics[width=\linewidth]{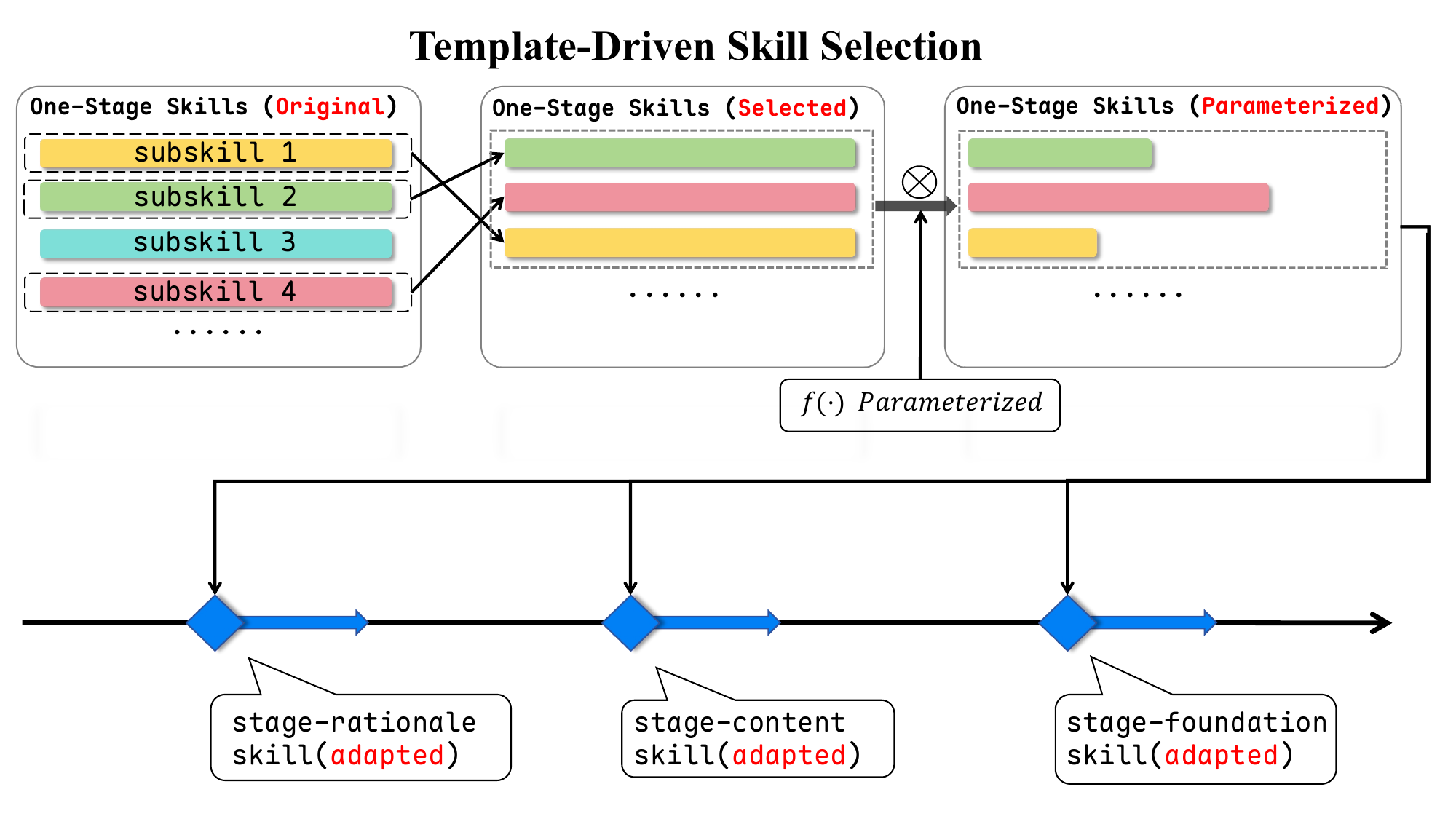}
\caption{Template-driven skill selection. Given a parsed template specification $T$, the system filters the full skill library into a compatible candidate set (Original $\to$ Selected), then applies the selection function $f(\cdot)$ with template-derived parameters $\theta_k$ (Parameterized) to produce three sets of adapted stage skills --- \texttt{stage-rationale}, \texttt{stage-content}, and \texttt{stage-foundation} --- deployed sequentially along the writing pipeline. Individual skills remain unchanged; only the assignment and parameters vary per template.}
\label{fig:template_driven}
\end{figure}

FundingClaw addresses this through a \textbf{Template Parser} that ingests the target agency's specification document and extracts its section hierarchy, per-section content, and page estimates into a structured template specification $T$. The pipeline then performs \textbf{template-driven skill selection} over the skill library $S = \{s_1, \dots, s_n\}$ and $S_k ~ (k \in \{1,2,3\})$ indicates the origin skill set before selected for each writing stage. Given the parsed template type $\tau(T)$, the system first \textit{filters} the applicable skills into a candidate set
\begin{equation}
  S_{\mathrm{cand}} = \{ s \in S \mid \mathrm{compatible}(s, \tau(T)) \},
\end{equation}
where $\mathrm{compatible}$ indicates that a skill can handle the template's section structure and writing language. A selection function $f$ then \textit{assigns}  skills to each of the three writing stages,
\begin{gather}
  f(S_k) = S'_k, \\
  S'_k = \{ s \mid s \in S_{\mathrm{cand}} \} \cap \{ s \mid s \in S_k \}, \\
  f_k(S_k) \in S_{\mathrm{cand}}, \\
  S^* = \{ S'_1, S'_2, S'_3 \}
\end{gather}
so the active skills are exactly the image $S^*$. Finally, each selected skill is \textit{parameterized} with template-derived settings,
\begin{equation}
  \theta_k = \mathrm{param}(T, S'_k),
\end{equation}
including the per-section length budgets and the citation scope allowed for that stage. Individual skills remain unchanged; only the assignment $f$ and the parameters $\theta_k$ vary per template. A Citation Scope Gate subsequently enforces the agency's citation policy as a static check over the generated files. The same research brief and skill library thereby produce structurally distinct, regulation-compliant outputs for NSFC, NSF, ERC, KAKENHI, or any agency whose template document is supplied, without modifying generation logic. The overall selection process is illustrated in Fig.~\ref{fig:template_driven}.

\subsection{Blueprint-Skeleton Aligned Serial Generation}
\label{sec:blueprint}

Grant proposals span 20--40 pages of tightly coupled content, yet LLMs generating long documents sequentially suffer from semantic drift across sections. Before any prose is written, FundingClaw derives a structured blueprint from the template specification and research brief, specifying section titles, key claim slots, length allocations, and figure placement markers. The three writing stages then execute sequentially; at the start of each, a Blueprint Alignment Check verifies that section structure, length allocations, and research framing remain consistent with prior stages, and triggers correction instructions to the incoming writer skills if drift is detected.

\subsection{Human Feedback Checkpoints}
At each stage boundary, the pipeline optionally pauses to expose the current blueprint and generated content to the researcher. The PI may issue natural-language correction instructions before the next stage resumes, enabling timely preference alignment without restarting the pipeline.

\subsection{Three-Stage Proposal Writing}
\label{sec:proposal_writing}

Writing a competitive grant proposal requires producing three qualitatively distinct bodies of content including a field-level motivation survey, a detailed technical plan, and a verifiable record of the PI's prior work and each demands different skills, different information sources, and different compliance rules. Dedicated stage-level subagents with targeted skill configurations are needed.

Under the selected skills and blueprint constraints, FundingClaw orchestrates three sequential stages to produce the full proposal body. Each stage follows a pattern: it first retrieves or verifies relevant evidence, then generates prose through a set of writer skills whose composition and length targets follow the template-derived parameters.

\paragraph{Research Rationale.}
The goal of this stage is to produce a literature-grounded argument establishing the significance and timeliness of the proposed research. LLMs prompted to write a literature survey directly tend to fabricate references or produce unsupported generalizations. FundingClaw addresses this with a two-phase design. A Literature Prefetch step assembles a candidate bibliography of $\geq$60 papers through structured web search queries, producing a per-direction literature memo. Two analytical skills then operate on this memo: the Significance Analyzer decomposes the motivation along a four-dimensional causal chain (phenomena, urgency, gaps, value), and the Landscape Analyzer identifies competing approaches and extracts concrete gap statements. Both produce memos containing only citation-anchored observations. Three writer skills (Significance, Survey, and Trends Writers) then consume these memos to produce narrative subsections, with every citation constrained to papers surfaced in the memos. This decoupling prevents citation fabrication and ensures coverage is driven by systematic literature examination. Once prose is complete, a Figure Generator reads the rationale text, extracts the two most visualizable concepts, and sends them to an image generation API; the returned figures are embedded directly into the LaTeX source, preventing figure-text inconsistency.

\paragraph{Research Content.}
This stage translates the research vision into a concrete technical plan across three interdependent dimensions. The Tasks Writer produces research tasks specifying inputs, methodology, expected outputs, evaluation metrics, and task dependencies. The Objectives Writer derives measurable milestones from the task definitions. The Questions Writer formulates 2--4 underlying scientific questions using a structured arc from observed phenomena through identified gaps to the proposed mechanism. Each writer receives the research brief, the blueprint, and the completed rationale findings as context. After generation, a Citation Scope Gate removes citation commands from sections where the agency's policy prohibits them.

\paragraph{Research Foundation.}
This stage establishes the PI's credibility to execute the proposed research. The central risk is factual error: misrepresenting publication records, citation counts, or author roles would damage the application's credibility. The system constructs a verified applicant profile through a three-source cross-validation protocol. The user-provided publication list is cross-referenced against Google Scholar, DBLP, and Semantic Scholar; each paper is annotated with its verified author role (first, co-first, or corresponding), citation count, and CCF venue ranking. Papers where the PI lacks a leading role are excluded from the listing. The verified publications are then partitioned into research directions by computing semantic similarity between each paper's content and the tasks defined in the previous stage, allowing the foundation section to mirror the proposal's own technical taxonomy rather than presenting a flat chronological list.

For agencies requiring detailed prior-work exposition, FundingClaw performs PDF-grounded content extraction. A multi-tier retrieval pipeline acquires each representative paper, and MinerU~\cite{wang2024mineruopensourcesolutionprecise} converts it to structured Markdown. The backbone model then extracts the core contribution, method, and findings under strict provenance constraints: every claim must trace to a specific passage, and no content may be generated from internal knowledge. The result is a set of figure-illustrated mini-reviews embedded in the proposal, replacing a bibliographic list with summaries of the PI's prior work.

\subsection{Quality Control and Compilation}
\label{sec:quality_control}

Automated long-form generation inevitably introduces hallucinations, internal inconsistencies, and format violations that no single writing pass can fully avoid. Without a dedicated review mechanism, these errors accumulate silently and can render an otherwise strong proposal non-compliant or scientifically unconvincing.

A Self-Review Agent evaluates the assembled draft across four dimensions: logical coherence, compliance, scientific merit, and feasibility. It assigns a score out of 10 and generates fix instructions; the pipeline applies fixes and re-reviews iteratively (up to 3 rounds for \texttt{balanced}, 8 for \texttt{beast}), with early exit at $\geq$8/10. The LaTeX Compiler then runs, and a Polish Agent performs language refinement before producing the final PDF.

\section{Experiments}
\label{sec:experiments}

\subsection{Experiment Setup}

\subsubsection{Dataset}

Evaluating an automated grant proposal writer is constrained by a fundamental data scarcity. In China, the United States, and most other countries, submitted grant proposals are confidential, and no public corpus exists to serve as reference answers. A competitive proposal is not uniquely determined by its topic, so no fixed reference template can serve as ground truth.

To obtain a practical evaluation setup, we adopt a topic-driven protocol. We collect \textbf{10 research topics} spanning contemporary AI and computing, each paired with a concise content summary in the style of a research direction statement, which serves as the generation input, ensuring evaluation is not overfit to a single sub-field. The complete list with full summaries is provided in Appendix~\ref{app:topics}.

\subsubsection{Evaluation Metrics}
\label{sec:eval_metrics}

\noindent\textbf{Multi-agency template rubrics.}~We evaluate proposals under four agency templates: \textbf{NSFC}, \textbf{NSF}, \textbf{ERC}, and \textbf{KAKENHI}. All four rubrics share the same eight dimensions (Table~\ref{tab:rubric_summary}), with per-agency point allocations adjusted to reflect each template's emphasis. We evaluate each type of proposal quality with a structured rubric comprising 78 items scored on a three-level scale (0, 0.5, 1) for a maximum of 78 points. The rubric dimensions and items were derived from the official review criteria of each funding agency and validated against experienced grant reviewers, not constructed to match FundingClaw's internal modules. The NSFC rubric is provided in Appendix~\ref{app:rubrics} as a representative example.

\noindent\textbf{LLM-judge evaluation.}~To assess self-preference bias, we score the same proposals with three independent judges (\textbf{Claude Sonnet 4.6}, \textbf{GPT-5.5}, and \textbf{DeepSeek V4 Flash}); results are reported in Section~\ref{sec:scoring_consistency}. All main experiments use \textbf{Claude Sonnet 4.6} as the unified judge.

\noindent\textbf{Image generation setup.}~All configurations use \textbf{Gemini 2.0 Flash Image} as the fixed image backend, so D6 measures each backbone's prompt-construction ability rather than the renderer.
           
\begin{table}[!t]
\centering\small\setlength{\tabcolsep}{4pt}
\caption{Summary of the eight evaluation dimensions of rubrics for NSFC.}
\label{tab:rubric_summary}
\begin{tabular}{clc}
\toprule
\textbf{Dim.} & \textbf{Focus} & \textbf{Pts} \\
\midrule
D1 & Rationale and scientific value & 13 \\
D2 & Research content and innovation & 16 \\
D3 & Plan and feasibility & 15 \\
D4 & Research foundation & 9 \\
D5 & Writing standards & 4 \\
D6 & Figure quality & 5 \\
D7 & Citation credibility & 10 \\
D8 & Internal consistency & 6 \\
\midrule
 & \textbf{Total} & \textbf{78} \\
\bottomrule
\end{tabular}
\end{table}
              
\subsubsection{Backbone}

FundingClaw supplies the writing workflow, and the underlying capability comes from whichever LLM drives it. We run every configuration with three backbones of \textbf{Claude Sonnet 4.6}, \textbf{GPT-5.5}, and \textbf{DeepSeek V4 Flash}. All backbones use default decoding settings and identical prompts and tool configurations.          

\subsubsection{Baselines and Our Pipeline}

We evaluate FundingClaw against two baselines on the same 10 topics.

\textbf{Simplified Pipeline.} Keeps the same template formatting, section-wise generation, and literature retrieval, but removes blueprint alignment, causal pre-analysis, and closed-loop self-refinement.

\textbf{ResearchAgent.} We adapt ResearchAgent~\cite{baek2024researchagent} as a research-ideation baseline: its pipeline produces structured \texttt{problem}, \texttt{method}, and \texttt{experiment} fields from the research brief, which are then rewritten into proposal sections.

\textbf{FundingClaw.} Our full pipeline: selects and orchestrates writing skills from the target template, runs causal pre-analysis, and generates the proposal stage by stage under blueprint alignment and closed-loop self-refinement.
              
\subsection{Self-Preference Bias and Scoring Consistency}
\label{sec:scoring_consistency}

To rule out self-preference bias, we score NSFC proposals from three backbones with all three judges under the same rubric (Table~\ref{tab:judge_consistency}). Rankings align closely across judges with no systematic self-inflation, so we fix Claude Sonnet 4.6 as the unified judge for all experiments.                  
                        
\begin{table}[t]
\centering\small\setlength{\tabcolsep}{3pt}
\caption{Cross-judge scoring consistency. Each cell reports the mean total scoring rate assigned by the row judge to proposals generated by the column backbone.}
\label{tab:judge_consistency}
\begin{tabular}{lccc}
\toprule
\textbf{Judge} $\backslash$ \textbf{Generator} & \textbf{Sonnet 4.6} & \textbf{GPT-5.5} & \textbf{DeepSeek} \\
\midrule
Claude Sonnet 4.6  & 69.20\% & 51.92\% & 68.40\% \\
GPT-5.5            & 68.42\% & 48.21\% & 67.37\% \\
DeepSeek V4 Flash  & 74.15\% & 52.60\% & 71.16\% \\
\bottomrule
\end{tabular}
\end{table}
 
\subsection{Human--LLM Judge Agreement}
\label{sec:human_eval}

\begin{figure}[t]
\centering
\includegraphics[width=0.75\columnwidth]{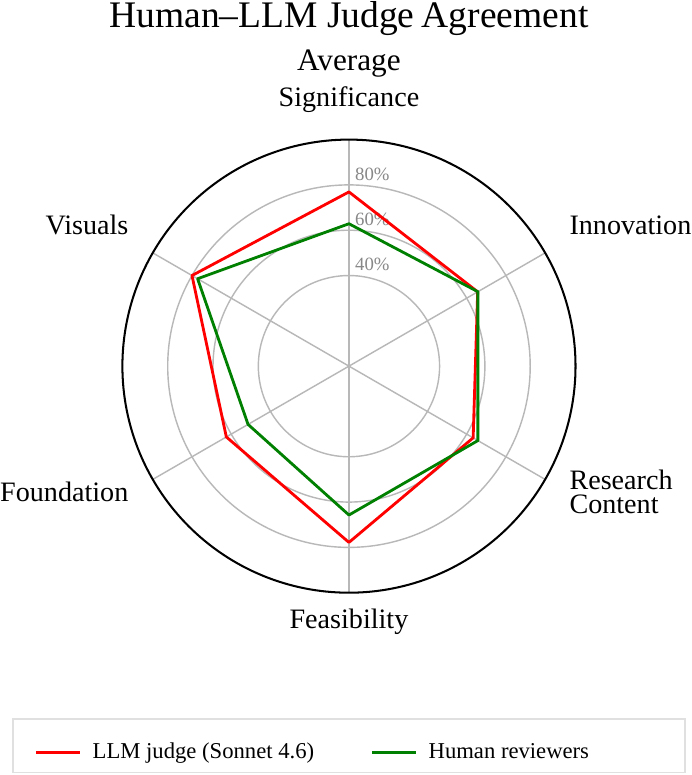}
\caption{Human--LLM judge agreement on proposals, shown as scoring rates over the six shared rubric dimensions. Human scores are averaged over all reviewers and normalised from the 1--5 scale.}
\label{fig:human_llm_agree}
\end{figure}

We validate the automatic judge against human expert scores. Experienced NSFC applicants and reviewers scored the same proposals on a six-dimension rubric (see Appendix~\ref{app:human_eval}); per-dimension rates are shown in Fig.~\ref{fig:human_llm_agree}. The two evaluators agree closely across most dimensions, validating the LLM judge for expert-level assessment.
                   
\subsection{Main Results}
\label{sec:main_results}

\begin{figure}[t]
\centering
\includegraphics[width=0.63\columnwidth]{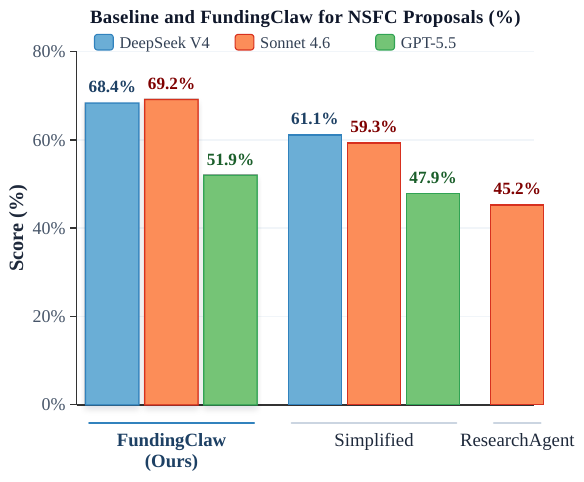}
\caption{Mean total scores on the NSFC template across all systems, grouped by backbone model.}
\label{fig:nsfc_results}
\end{figure}

FundingClaw outperforms both baselines across all backbone models (Fig.~\ref{fig:nsfc_results}). The gap over ResearchAgent is substantial: Sonnet 4.6 reaches 69.2\% versus ResearchAgent's 45.2\%, a difference of nearly 24 percentage points. FundingClaw also outperforms the Simplified pipeline by 7--10 pp across all three backbones, confirming that blueprint alignment, causal pre-analysis, and closed-loop self-refinement each contribute meaningfully to proposal quality.
   
\begin{figure}[t]
\centering
\includegraphics[width=\columnwidth]{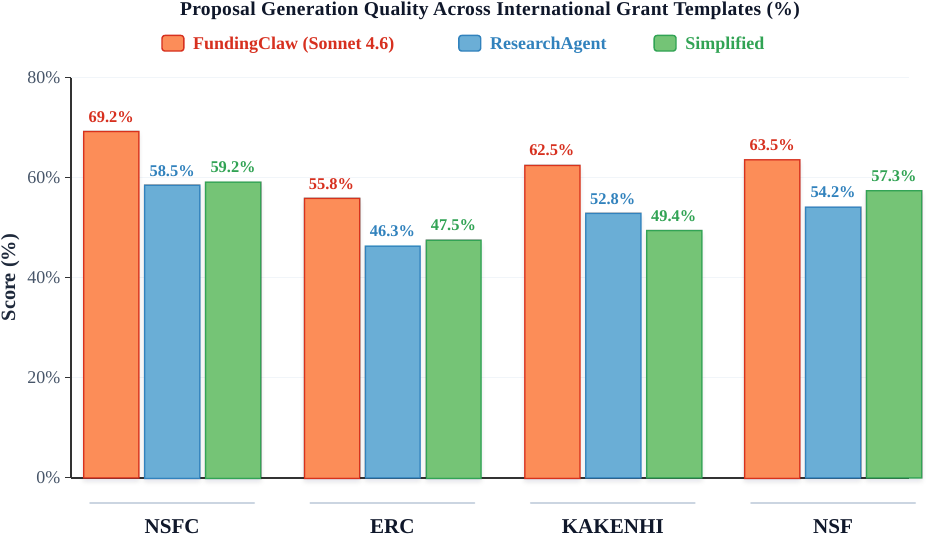}
\caption{Pipeline scores across four international grant templates by backbone model.}
\label{fig:template_results}
\end{figure}

\noindent\textbf{Multi-Template Generalization.}~Fig.~\ref{fig:template_results} shows that FundingClaw's advantage holds consistently across all four agency templates, confirming that template-driven skill selection generalises beyond the NSFC setting rather than overfitting to a single rubric. 
        
\begin{figure}[t]
\centering
\includegraphics[width=\columnwidth]{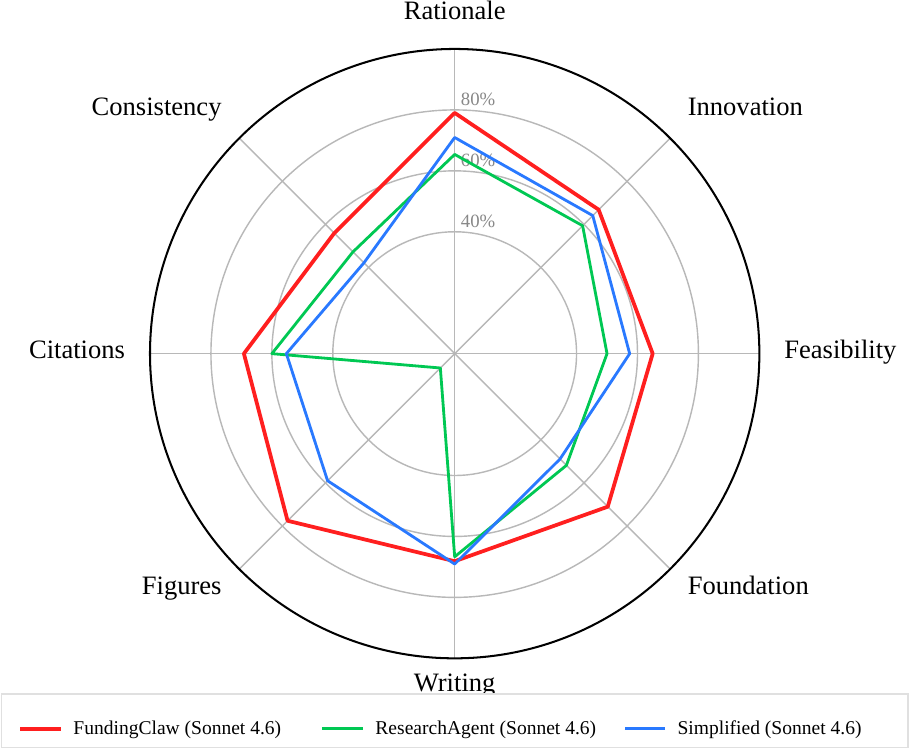}
\caption{Per-dimension scoring profile on the NSFC rubric with Sonnet 4.6, normalised to each dimension's maximum.}
\label{fig:system_radar}
\end{figure}

\noindent\textbf{Dimension-level comparison.}~FundingClaw's advantage is most pronounced on the research rationale dimensions including foundation, citation credibility, and rationale (Fig.~\ref{fig:system_radar}) which reviewers scrutinise most heavily. These dimensions demand deep literature grounding, multi-source applicant verification, and causal motivation, all directly addressed by FundingClaw's pre-analysis stages.
   
\subsection{Ablation Study}
\label{sec:ablation}                                     
                      
To quantify the independent contribution of each component, we evaluate ten conditions that each remove or replace one element of the full pipeline, scoring all outputs with the same eight-dimension rubric. Results are shown in Fig.~\ref{fig:ablation}; full condition descriptions and full per-condition analysis is in Appendix~\ref{app:ablation}.

\textbf{Significance analysis matters most.} Removing significance analysis (C2) produces the single largest score drop, collapsing the rationale dimension most severely. The four-dimensional causal chain inlucding phenomena, urgency, gaps, value cannot be reconstructed by the writing stage from raw literature alone; without it, scientific motivation becomes assertion rather than argument.

\textbf{Literature retrieval cascades through the pipeline.} Removing the literature prefetch (C1) degrades citation credibility, rationale depth, and plan coherence simultaneously, since every analytical and writing skill draws from the same literature memo. Literature grounding is the shared substrate that makes structured pre-analysis possible.
   
\textbf{Figure scheduling is a dedicated component; logical coherence governs figure quality indirectly.} Removing C5 zeroes the figure-quality dimension and depresses feasibility scores. The remaining ablations (C3, C4, C6--C8) show smaller but consistent degradation; components such as blueprint alignment (C3) and self-refinement (C4) govern internal consistency, and when absent, the figure prompt inherits incoherent content, degrading the generated diagram even though the scheduler itself is intact. 

\begin{figure}[t]
\centering
\includegraphics[width=\columnwidth]{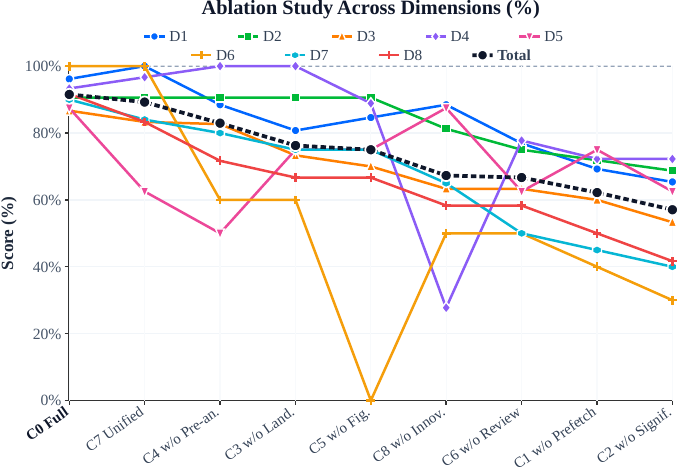}
\caption{Ablation study results. Each condition removes or replaces one component of FundingClaw (C0). Scores are total points out of 78.}
\label{fig:ablation}
\end{figure}
\section{Conclusion}
\label{sec:conclusion}

We introduced \textbf{FundingClaw} for automated grant proposal writing across heterogeneous funding templates. Template-driven skill selection reduces agency adaptation to a data-driven selection task; blueprint-skeleton aligned serial generation counters semantic drift. Three sequential stages cover the rationale, research content, and research foundation, with optional human checkpoints and a closed-loop self-review. Evaluated with an eight-dimension rubric across four agency templates, FundingClaw outperforms the baselines and ablations confirm each component contributes meaningfully. To our knowledge, it is the first system generating template-compliant proposals across multiple funding agencies. FundingClaw lowers the barrier for researchers navigating complex grant requirements and opens a path toward more systematic, evidence-grounded automated proposal writing.

\section*{Limitations and Future Work}

While FundingClaw demonstrates promising results in automatically generating grant proposals, there remain several limitations. First, the pipeline is built around the conventional proposal structure that contains research-grounded rationale, content, and foundation sections; for unconventional grant applications whose templates deviate substantially from this structure, the current skill selection may not adapt flexibly enough to produce compliant output directly. Extending coverage to a broader range of agency formats would enhance generality. Second, generation is time-consuming and involves many sequential steps, with a relatively high token consumption and corresponding computational cost. Future work could investigate lightweight routing and more efficient token usage to reduce generation overhead. Finally, due to the confidentiality of real grant applications, publicly accessible proposal datasets that could serve as reference are scarce, which constrains both training and benchmarking. We plan to construct a public benchmark from synthetic and de-identified proposals while validating our scores against human expert review panels, which would provide a more rigorous foundation for future research in this area.

\section*{Broader Impact and Ethics Statement}

FundingClaw is designed to reduce the heavy writing burden on PIs by automating the labor-intensive processes of literature synthesis, document structuring, and compliance formatting, thereby freeing researchers to focus on the scientific substance of their work. The system is an authoring assistant, not an autonomous submitter: human feedback checkpoints at each pipeline stage require the PI to review and correct generated content, and researchers remain fully responsible for the scientific accuracy and integrity of the final proposal.

\noindent\textbf{Data and privacy.}~
The development of FundingClaw did not involve any private or sensitive data. The grant proposals used as references during development were provided by collaborators with their explicit consent, and no confidential submissions from third parties were used. The ten evaluation topics used in our benchmark were derived from de-identified research direction summaries, containing no personally identifiable information or proprietary content.

\noindent\textbf{Confidentiality of proposal materials.}~
Grant proposals may contain sensitive information including unpublished research findings, proprietary data, trade secrets, and patent-pending material. FundingClaw's pipeline addresses generation workflow only and does not itself introduce data-leakage risk beyond what is inherent in the underlying model calls. When cloud-hosted backbone models are used, data handling is governed by the respective provider's terms of service, and users should review those agreements before processing confidential content. For projects with strict confidentiality requirements, we recommend deploying FundingClaw with a locally hosted open-source model to ensure that sensitive proposal materials do not leave the user's computing environment.

\noindent\textbf{Intended use and misuse prevention.}~
The primary goal of FundingClaw is to lower the writing barrier for researchers, particularly those without access to professional grant-writing support, and is not intended to facilitate academic misconduct. The applicant-record verification module cross-validates publication claims against public databases and excludes papers where the PI's leading authorship role cannot be confirmed, reducing the risk of credential misrepresentation. We encourage users to disclose the use of AI-assisted writing tools in accordance with the policies of their target funding agency.

\noindent\textbf{Use of AI assistance.}~
In preparing this paper, we used Claude Code and Claude Sonnet 4.6 to polish language and correct grammatical errors. All scientific content, experimental design, results, and conclusions are the work of the authors.

\bibliography{custom}

\appendix
\twocolumn
\section{Evaluation Topics}
    \label{app:topics}

    The complete list of the 10 evaluation topics with their full content summaries is provided below.

    \paragraph{T1: Memory Design for Multimodal Video Large Models.}
    \textbf{Title:} Research on Spatiotemporal Memory Mechanisms and Dynamic Updating for Streaming Multimodal Large Models in Ultra-Long Video Understanding.

    \textbf{Content Summary:} Currently, multimodal video large models such as Sora face critical bottlenecks when processing ultra-long videos, including limited context windows, quadratic computational complexity explosion, and catastrophic forgetting. This project proposes to study a bionic, hierarchical spatiotemporal memory architecture that decouples short-term working memory from long-term episodic memory. By introducing dynamic reading and writing, memory compression, and semantic-based retrieval mechanisms, the model aims to achieve continuous understanding and spatiotemporally consistent reasoning over ultra-long video streams without recomputing global attention. This research plays a key role in surmounting the GPU memory and computational limits of long video generation, complex video QA, and real-time streaming media analysis.

    \paragraph{T2: Instruction Following in Action-Conditioned World Models.}
    \textbf{Title:} Research on Multi-Granularity Instruction Alignment and Precise Following for Action-Conditioned World Models under Physical Law Constraints.

    \textbf{Content Summary:} Action-conditioned world models hold great potential for trajectory prediction in embodied AI. However, current models often suffer from semantic understanding drift and physical hallucinations, i.e., generating future states that violate physical laws, when executing complex natural language instructions, which remains a core bottleneck for practical deployment. This project aims to investigate the cross-modal alignment mechanism from abstract linguistic instructions to fine-grained physical action spaces. By incorporating physical engine priors and an instruction-following reward model based on human feedback, the project aims to constrain the world model to strictly follow instruction semantics and dynamical laws when generating future states, thereby significantly enhancing the generalization and execution accuracy of embodied agents under open-ended instructions.

    \paragraph{T3: Closed-Loop Simulation for Autonomous Driving Based on 4DGS.}
    \textbf{Title:} High-Fidelity Closed-Loop Simulation and Generalization Verification for End-to-End Autonomous Driving Based on Interactive 4D Gaussian Splatting.

    \textbf{Content Summary:} The training of end-to-end autonomous driving models relies heavily on massive amounts of data. However, traditional simulators lack realism, and real-world road tests struggle to cover corner cases. While 4D Gaussian Splatting can achieve highly realistic scene reconstruction, its static, read-only nature makes dynamic interaction difficult, presenting a bottleneck for closed-loop simulation. This project proposes to study 4DGS-based scene decoupling and physical-level interactive editing techniques for dynamic objects such as vehicles and pedestrians. It aims to build a closed-loop simulation engine that supports real-time rendering and physical-level sensor simulation, providing a high-fidelity, interactive testing and reinforcement learning environment for end-to-end autonomous driving large models.

    \paragraph{T4: Error Correction Capability in Embodied AI Models.}
    \textbf{Title:} Research on Multimodal Perceptual Feedback and Autonomous Closed-Loop Error Correction Mechanisms for Embodied Large Models in Open Environments.

    \textbf{Content Summary:} When executing tasks in the physical world, embodied agents are highly prone to error accumulation caused by environmental disturbances or planning deviations. Current open-loop execution strategies lack self-reflection and error-correction capabilities, which is a major pain point preventing embodied models from practical application. This project proposes to research real-time execution monitoring and anomaly detection technology based on Vision-Language Models. By integrating visual, tactile, and proprioceptive feedback, the study aims to construct a closed-loop cognitive architecture of planning, execution, evaluation, reflection, and replanning. This will enable agents to autonomously analyze the causes of action failures and dynamically generate corrective strategies, fundamentally improving the robustness of embodied models in complex, unstructured environments.

    \paragraph{T5: EEG-Based Epilepsy Seizure Prediction.}
    \textbf{Title:} Research on Cross-Subject Generalized EEG Spatiotemporal Representation Foundation Models and Early Warning Mechanisms for Epileptic Seizures.

    \textbf{Content Summary:} EEG-based epilepsy prediction is critical to patient health and life, but the extremely low signal-to-noise ratio of EEG signals and significant individual differences severely limit the accuracy and real-time performance of prediction models. This project plans to introduce self-supervised learning and masked autoencoder architectures to pre-train an EEG spatiotemporal representation foundation model using massive unlabelled EEG data. Combined with lightweight, few-shot adaptation techniques, the project aims to extract highly robust pre-ictal seizure precursor features, with the goal of realizing a high-precision, low-latency, wearable early warning system for epileptic seizures that can rapidly adapt to new patients.

    \paragraph{T6: Enhancing Long-Horizon Task Capabilities in Embodied AI.}
    \textbf{Title:} Research on Autonomous Planning and Execution of Long-Horizon Complex Tasks in Embodied AI Driven by Hierarchical Language-Action Abstractions.

    \textbf{Content Summary:} When handling long-horizon tasks, embodied AI faces the bottlenecks of action space dimension explosion and extremely sparse rewards, which make traditional reinforcement learning difficult to converge. This project proposes a hierarchical control architecture based on the collaboration between Large Language Models and low-level action policies. The upper layer leverages LLMs for high-level semantic decomposition and common-sense reasoning, breaking long-horizon tasks down into short-range sub-goals, while the lower layer utilizes parameterized skill libraries to execute atomic actions. The research will focus on the verification of state transitions between sub-goals and skill chaining mechanisms, seeking to push past the execution limits of embodied agents in long-cycle tasks within complex scenarios such as kitchens and factories.

    \paragraph{T7: Enhancing 3D Perception in Embodied Action Models via Visual Geometry Foundation Models.}
    \textbf{Title:} Research on 3D Spatial Perception and Interaction Affordance for Embodied Action Policies Integrating Visual Geometry Foundation Models.

    \textbf{Content Summary:} Current mainstream Vision-Language-Action models in embodied AI mostly rely on 2D image features and lack a deep understanding of depth, occlusion, and 3D spatial geometry, resulting in low success rates in complex grasping and fine manipulation. This project proposes to research the integration of powerful 3D prior knowledge from visual geometry foundation models such as VGGT into embodied action policies. By extracting the 3D topological structure and interaction affordance of scenes, the project aims to address the ambiguity in 2D-to-3D mapping, thereby significantly improving the 3D spatial reasoning capabilities of embodied models under single or few views and their manipulation generalization across various object morphologies.

    \paragraph{T8: Efficient Training and Inference of Embodied World Action Models.}
    \textbf{Title:} Research on Lightweight Architectures and Efficient Edge Inference for Embodied World Action Models Oriented towards Real-Time Control.

    \textbf{Content Summary:} Embodied World Action Models have shown immense potential by predicting future states to guide decision-making. However, WAMs based on Transformers or diffusion models suffer from high computational complexity and severe inference latency bottlenecks, making them unable to meet the demands of high-frequency real-time robotic control. This project aims to explore a next-generation WAM architecture based on State Space Models such as Mamba or Latent Consistency Models. It will investigate linear-complexity modeling of temporal features and autoregressive decoding acceleration techniques, aiming to achieve low-latency real-time inference on robot edge hardware while maintaining prediction accuracy.

    \paragraph{T9: Social Loafing in Multi-Agent Collaboration.}
    \textbf{Title:} Research on the Evolution Mechanism and Dynamic Incentive Intervention of Social Loafing in Large Model-Driven Multi-Agent Collaboration.

    \textbf{Content Summary:} In multi-agent reinforcement learning and LLM-based collaborative systems, the ambiguity of global reward distribution often causes agents to evolve free-riding behaviors or a laziness effect, i.e., social loafing. This leads the overall system efficiency to fall into sub-optimal equilibria, presenting a major bottleneck to the emergence of swarm intelligence. This project proposes to combine game theory and multi-agent reinforcement learning to establish mathematical formulations and evolutionary dynamics models of the social loafing effect. It intends to develop fine-grained credit assignment mechanisms based on dynamic Shapley values along with intrinsic motivation reward designs, aiming to structurally eliminate lazy behavior and incentivize active cooperation among heterogeneous agents.

    \paragraph{T10: Efficient Real-Robot Reinforcement Learning in Embodied AI.}
    \textbf{Title:} Research on Low-Sample Real-Robot Online Reinforcement Learning Mechanisms for Embodied AI under Sim-to-Real Transfer and Safety Constraints.

    \textbf{Content Summary:} Training embodied AI solely in simulation faces a severe sim-to-real gap, whereas training reinforcement learning directly on physical robots is constrained by extremely low sample efficiency and high hardware trial-and-error costs. This project proposes to research an efficient real-robot learning paradigm of offline pre-training plus safe online fine-tuning. By constructing conservative offline policies in simulation, combined with epistemic uncertainty estimation and Control Barrier Functions during real-world exploration, the project aims to enable rapid closed-loop adaptive policy evolution using minimal real physical interaction data, while ensuring the absolute safety of the robotic hardware.

    \clearpage
    \onecolumn
    \section{Ablation Details}
    \label{app:ablation}
    \setlength{\floatsep}{4pt}
    \setlength{\textfloatsep}{8pt}

    Table~\ref{tab:ablation_components} lists each ablation condition and the function of the removed or replaced component. The remaining conditions are described here for completeness.

    \begin{table}[h!]
    \centering\renewcommand{\arraystretch}{0.85}
    \caption{Component functions and the primary rubrics affected by each ablation.}
    \label{tab:ablation_components}
    \resizebox{\textwidth}{!}{
    \begin{tabular}{lp{0.72\textwidth}}
    \toprule
    \textbf{Condition} & \textbf{Function of the removed or replaced component} \\
    \midrule
    C1 w/o Literature Prefetch & Assembles a candidate bibliography of at least 60 papers by issuing structured web search queries derived from the research brief, producing a per-direction literature memo that feeds all analytical and writing skills. \\
    C2 w/o Significance Analysis & Decomposes the research motivation along a four-dimensional causal chain from observed phenomena, practical urgency, gaps in existing approaches, to the value of further study, producing a significance memo containing only citation-anchored factual observations. \\
    C3 w/o Landscape Analysis & Maps competing research groups, constructs a claims-evidence matrix, and extracts concrete gap statements from the retrieved literature, grounding the survey in a structured view of the field. \\
    C4 w/o All Pre-analysis (C2+C3) & Removes both the significance and the landscape analysis together, leaving only the raw prefetched literature memo without any structured analytical layer. \\
    C5 w/o Figures & Generates the motivation and workflow figures and embeds them into the rationale and content sections, providing the visual argument that supports the written claims. \\
    C6 w/o Auto-Review & Runs the closed-loop self-review that evaluates the draft across logical coherence, compliance, scientific merit, and feasibility, then applies iterative fixes before compilation. \\
    C7 Unified Rationale & Replaces the three-section rationale split with a single merged section, removing the explicit separation of significance, survey, and trends. \\
    C8 w/o Innovation Section & Writes the dedicated innovation points subsection that lists the project's distinctive characteristics and novelty relative to existing work. \\
    \bottomrule
    \end{tabular}
    }
    \end{table}

    \paragraph{C1: Literature Prefetch.}
    Removing the literature prefetch forces analytical and writing skills to operate without a shared literature memo, so each skill must independently reconstruct field context from scratch. This produces systematic degradation across three dimensions: citation credibility (D7) suffers most directly as citations are no longer drawn from a pre-verified bibliography; rationale depth (D1) declines because gap analysis cannot be anchored to concrete paper evidence; and plan coherence (D3) weakens because task decomposition can no longer be grounded in a mapped landscape of existing approaches. The simultaneous degradation across all three reflects that the literature memo is the shared substrate that makes structured pre-analysis possible---without it, the pipeline proceeds on internal LLM knowledge alone, which hallucinates citations and misrepresents the state of the art.

    \paragraph{C2: Significance Analysis.}
    The significance analyzer decomposes research motivation into a four-dimensional causal chain: phenomena (what is observed), urgency (why it matters now), gaps (what approaches have been tried and why they fail), and value (what solving the problem enables). Removing this component (C2) produces the single largest score drop across all conditions. The rationale dimension (D1) collapses most severely because the causal chain that drives a compelling scientific argument cannot be reconstructed by the writing stage from raw literature alone---the writing skill receives the literature memo but lacks the structured analytical layer needed to synthesize it into a four-part motivation. Without significance analysis, scientific motivation reverts to assertion rather than argument, and the proposal fails the gap-innovation traceability checks in D2.

    \paragraph{C3: Landscape Analysis.}
    The landscape analyzer maps competing approaches, constructs a claims-evidence matrix, and extracts concrete gap statements from the retrieved literature, grounding the survey in a structured view of the field. Without C3, the survey writer lacks explicit gap statements and must invent them from raw memo content. This produces a shallower literature survey (D1.2) and weakens innovation defensibility (D2.1), as the claimed novelty is no longer contrasted against a structured view of competing work. The degradation from C3 removal is smaller than C2 because the significance analyzer already extracts partial gap information; C3 provides complementary breadth over the landscape rather than causal depth.

    \paragraph{C4: Combined Pre-analysis Removal (C2+C3).}
    C4 removes both significance and landscape analysis simultaneously. Interestingly, C4 yields a higher overall score than either C2 or C3 alone. This occurs because retaining only one analytical component produces asymmetric intermediate artifacts: a significance memo without landscape context, or landscape gaps without causal framing, creates partial analytical structure that misleads downstream writing skills into building cross-section arguments on an incomplete foundation, producing traceability failures in D2 and D8. When both modules are removed together, the writing stage falls back to direct drafting from the raw literature list, bypassing the intermediate logical mismatches---it lacks the causal and competitive depth that C2 and C3 provide, but it avoids the additional degradation caused by half-structured analytical input.

    \paragraph{C5: Figure Generation.}
    Removing C5 zeroes the figure-quality dimension (D6) and depresses feasibility scores (D3). The figure generator reads the completed rationale text, identifies the two most visualizable concepts, and sends prompts to an image generation API; the returned figures are embedded directly in the LaTeX source. Without this step no figures are present, causing D6 items R6.1--R6.5 to all score 0 under the rubric's no-figure pre-check rule, and the plan and feasibility dimension loses the visual roadmap and framework figures that D3.3 evaluates. The figure scheduler is a dedicated, isolated component: its removal produces no indirect effect on the written content, which is why C5 degrades D6 fully while leaving text-based dimensions largely intact. By contrast, ablations such as C3 and C4 indirectly degrade D6 because their absence causes the written content passed to the figure generator to be less coherent, producing lower-quality prompts even though the scheduler itself is intact.

    \paragraph{C6: Self-Review Agent.}
    The self-review agent evaluates the assembled draft for logical coherence, compliance, scientific merit, and feasibility, then applies iterative fixes (up to 3 rounds in balanced mode, 8 in beast mode) with early exit at a score of 8/10. Removing C6 produces consistent but moderate degradation across the internal consistency dimension (D8) and the writing standards dimension (D5). The agent's primary role is error correction rather than content generation, so proposals without self-review still contain all required sections but accumulate unresolved hallucinations, inconsistencies between sections, and format violations that a single-pass generation cannot fully avoid.

    \paragraph{C7: Unified Rationale.}
    C7 replaces the three-section rationale structure (significance, survey, and trends writers producing separate subsections) with a single merged rationale section. Removing the explicit split degrades coverage of individual sub-dimensions in D1: the significance subsection (D1.3), the literature survey (D1.2), and the trends discussion each receive less focused attention when produced under a single undifferentiated prompt. This confirms that the three-section decomposition serves a functional role: it forces the writing stage to produce complete coverage of each sub-requirement rather than distributing effort unevenly across a merged output.

    \paragraph{C8: Innovation Section.}
    C8 removes the dedicated innovation points subsection that lists the project's distinctive characteristics and novelty relative to existing work. This directly degrades the innovation depth items in D2.1 (R2.1.1--R2.1.3), which require itemized, technically specific innovations to receive full credit. The proposal's innovations are still present implicitly in the research content, but without an explicit innovation section the rubric cannot trace them to named gap-innovation mappings, and the internal consistency dimension (D8) also weakens because the logic chain from gap to innovation to experimental verification becomes harder to follow. Critically, omitting the innovation section also triggers a catastrophic collapse in the research foundation dimension (D4 plunges to 27.8\%). Under the evaluation rubric (R4.1.2 and R4.1.5), applicant track records and foundational papers must be explicitly mapped to named, itemized innovations; without an explicit innovation section, the evaluator fails this cross-section mapping, dropping D4.1 scores to zero.

    \clearpage
    \onecolumn
    \raggedbottom
    \section{Evaluation Rubrics}
    \label{app:rubrics}

    The complete evaluation rubric consists of eight dimensions and 78 scoring items. Each item is scored on a three-level scale, 0 for absent or fundamentally deficient, 0.5 for present but shallow, and 1 for present with depth and quality. The maximum total score is 78.

    \subsection{D1: Rationale and Scientific Value (13 pts)}
    \begin{notebox}{D1: Rationale and Scientific Value}
    This dimension evaluates the necessity, scientific value, and quality of problem formulation in the research.

    \noindent\textbf{D1.1 Gap Analysis Quality (5 pts).}
    \begin{itemize}\setlength{\itemsep}{0pt}\setlength{\parsep}{0pt}\setlength{\topsep}{0pt}
        \item \textbf{R1.1.1.} Are the identified research gaps supported by systematic literature analysis with explicit comparison of at least 3 technical approaches or methodological schools? Scoring: 0, no comparative analysis or fewer than 3 approaches; 0.5, lists approaches but no critical comparison; 1, deep comparative analysis showing why each approach is insufficient.
        \item \textbf{R1.1.2.} Are the research gaps grounded in recent high-quality references, with specific technical limitations traced to concrete evidence in cited papers? Scoring: 0, gaps asserted without evidence; 0.5, general citations but no specific technical limitations extracted; 1, each gap supported by specific findings from recent papers published in 2023 or later.
        \item \textbf{R1.1.3.} Does the gap analysis explain the fundamental reasons why existing methods fail, rather than merely describing phenomena? Scoring: 0, only describes what is missing; 0.5, surface-level why; 1, deep causal analysis covering theoretical limits, computational barriers, and data constraints.
        \item \textbf{R1.1.4.} Are quantitative claims about existing methods traceable to specific cited references with no unverifiable statistics? Scoring: 0, fabricated or unverifiable claims present; 0.5, all technical performance claims have citations but at most 2 market or macro statistics lack sources; 1, all quantitative claims traceable to specific references. Any technical performance figure without a citation reference number drops this item to 0 regardless of other claims.
        \item \textbf{R1.1.5.} Does the rationale present a clear scientific problem statement that synthesizes the identified gaps into a coherent research challenge? Scoring: 0, no synthesis or problem statement; 0.5, generic problem statement; 1, precise, bounded scientific problem derived from gap analysis.
    \end{itemize}

    \noindent\textbf{D1.2 Literature Coverage and Currency (4 pts).}
    \begin{itemize}\setlength{\itemsep}{0pt}\setlength{\parsep}{0pt}\setlength{\topsep}{0pt}
        \item \textbf{R1.2.1.} Does it cite at least 15 high-quality journal or conference papers directly related to the core research problem? Scoring: 0, fewer than 10 papers; 0.5, 10 to 14 papers or at least 15 with weak relevance; 1, at least 15 papers with strong topical relevance.
        \item \textbf{R1.2.2.} Does the reference list include at least 15 papers published in the last 3 years? Scoring: 0, fewer than 10 recent papers; 0.5, 10 to 14 recent papers; 1, at least 15 recent papers.
        \item \textbf{R1.2.3.} Does it provide a systematic review covering at least 3 major research directions or methodological schools with clear organization? Scoring: 0, scattered references with no structure; 0.5, lists 3 or more directions but no synthesis; 1, well-organized review showing the research landscape.
        \item \textbf{R1.2.4.} Does it comparatively analyze domestic and international research progress, identifying specific differences or regional gaps? Scoring: 0, no comparison or a single generic sentence; 0.5, at least 3 sentences dedicated to domestic research teams or institutions by name with specific characterization; 1, explicit side-by-side comparison identifying concrete technical differences between domestic and international state of the art.
    \end{itemize}

    \noindent\textbf{D1.3 Significance and Motivation (3 pts).}
    \begin{itemize}\setlength{\itemsep}{0pt}\setlength{\parsep}{0pt}\setlength{\topsep}{0pt}
        \item \textbf{R1.3.1.} Does it explicitly articulate the theoretical significance with expected contributions to foundational scientific knowledge? Scoring: 0, none stated; 0.5, generic claims; 1, specific theoretical contributions described.
        \item \textbf{R1.3.2.} Does it explicitly articulate the applied value with potential impact on engineering practice or industry? Scoring: 0, none stated; 0.5, generic impact claims; 1, concrete application scenarios with impact analysis.
        \item \textbf{R1.3.3.} Does it explain the relevance to national S\&T strategic directions or field development trends? Scoring: 0, no strategic relevance; 0.5, mentions strategy generically; 1, explicit connection to specific national priorities.
    \end{itemize}

    \noindent\textbf{D1.4 Visual Motivation (1 pt).}
    \begin{itemize}\setlength{\itemsep}{0pt}\setlength{\parsep}{0pt}\setlength{\topsep}{0pt}
        \item \textbf{R1.4.1.} Does it visually present, via a Motivation-type figure, the contrast between existing method deficiencies and this research's entry point? Scoring: 0, no motivation figure; 0.5, figure present but unclear contrast; 1, clear visual argument showing gap and approach. This item requires an actual embedded figure and scores 0 if no figures are embedded.
    \end{itemize}
    \end{notebox}

    \subsection{D2: Research Content and Innovation Assessment (16 pts)}
    \begin{notebox}{D2: Research Content and Innovation Assessment}
    This dimension evaluates whether research objectives are clear, content is reasonably decomposed, innovations are well-founded, and logical correspondence with the rationale is established.

    \noindent\textbf{D2.1 Innovation Depth and Novelty (7 pts).}
    \begin{itemize}\setlength{\itemsep}{0pt}\setlength{\parsep}{0pt}\setlength{\topsep}{0pt}
        \item \textbf{R2.1.1.} Are 2 to 4 innovations explicitly listed with technical specificity? Scoring: 0, no itemized innovations or fewer than 2; 0.5, listed but generic descriptions; 1, specific technical innovations with implementation detail.
        \item \textbf{R2.1.2.} Can each innovation be directly linked to a specific research gap from the rationale via keyword and concept tracing? Scoring: 0, no traceability; 0.5, weak correspondence; 1, clear one-to-one gap-innovation mapping.
        \item \textbf{R2.1.3.} Are the claimed innovations distinguishable from existing literature? Scoring: 0, identical to cited work or no comparison made; 0.5, explicit named comparison to the most similar published work with sufficient differentiation details; 1, clear novelty with named distinction from the closest prior art, supported by technical specifics. Innovations described only as proposals without explicit comparison to the closest existing work receive 0.
        \item \textbf{R2.1.4.} Do the innovations address fundamental technical challenges rather than engineering conveniences? Scoring: 0, only incremental improvements; 0.5, mix of incremental and substantive; 1, all innovations target core scientific barriers.
        \item \textbf{R2.1.5.} Are the innovations technically feasible given the proposed method and team capability? Scoring: 0, unrealistic or unsupported claims; 0.5, at least one specific feasibility argument, such as a directly applicable published team result, a cited paper demonstrating technical feasibility, or confirmed named resource availability; 1, each innovation has concrete feasibility evidence.
        \item \textbf{R2.1.6.} Does a web search confirm the absence of highly similar published work prior to this proposal? Scoring: 0, search finds at least one highly similar paper that is neither cited nor distinguished; 0.5, related work exists and is cited and distinguished, or the novelty claim is hard to fully confirm; 1, search confirms no highly similar prior work at the method level for all innovations.
        \item \textbf{R2.1.7.} Do the innovations show sufficient technical grounding, such as formulas, complexity analysis, or pilot data, to justify the design choices? Scoring: 0, all innovations described only as proposals with no formal basis; 0.5, at least 2 innovations include a formula or complexity analysis; 1, every innovation includes formal derivation or pilot data causally supporting the design choice.
    \end{itemize}

    \noindent\textbf{D2.2 Research Objectives and Content Structure (5 pts).}
    \begin{itemize}\setlength{\itemsep}{0pt}\setlength{\parsep}{0pt}\setlength{\topsep}{0pt}
        \item \textbf{R2.2.1.} Does it state an overall research objective that is specific, measurable, and bounded? Scoring: 0, no objective or too vague; 0.5, stated but broad; 1, precise, quantifiable objective.
        \item \textbf{R2.2.2.} Is research content decomposed into 2 to 4 relatively independent and mutually supportive sub-directions with clear boundaries? Scoring: 0, no decomposition or more than 4; 0.5, decomposed but overlapping; 1, clear, modular sub-directions.
        \item \textbf{R2.2.3.} Does each sub-direction include clear sub-objectives and expected outcomes that are verifiable? Scoring: 0, missing for at least one; 0.5, present but generic; 1, specific, quantifiable outcomes for each.
        \item \textbf{R2.2.4.} Are at least 2 key scientific problems explicitly identified with clear technical challenges? Scoring: 0, fewer than 2 or too vague; 0.5, identified but not deeply characterized; 1, precise technical challenges with difficulty analysis.
        \item \textbf{R2.2.5.} Is the progressive relationship or technical dependency between sub-directions explained? Scoring: 0, no dependency analysis; 0.5, mentioned but unclear; 1, clear dependency graph showing technical flow.
    \end{itemize}

    \noindent\textbf{D2.3 Logic Chain from Rationale to Content (2 pts).}
    \begin{itemize}\setlength{\itemsep}{0pt}\setlength{\parsep}{0pt}\setlength{\topsep}{0pt}
        \item \textbf{R2.3.1.} Can each sub-direction be traced to at least one specific research gap from the rationale? Scoring: 0, no traceability; 0.5, partial correspondence; 1, complete gap-to-content mapping.
        \item \textbf{R2.3.2.} Are quantitative improvement targets grounded in baseline values from the rationale or cited references? Scoring: 0, floating targets without baseline; 0.5, some baselines present; 1, all targets anchored to stated baselines.
    \end{itemize}

    \noindent\textbf{D2.4 Scale and Ambition (2 pts).}
    \begin{itemize}\setlength{\itemsep}{0pt}\setlength{\parsep}{0pt}\setlength{\topsep}{0pt}
        \item \textbf{R2.4.1.} Is the overall research scope commensurate with a 4-year NSFC General Program? Scoring: 0, clearly mismatched; 0.5, borderline appropriate; 1, well-calibrated to the funding cycle.
        \item \textbf{R2.4.2.} Is the temporal ordering in the annual plan consistent with technical dependencies? Scoring: 0, clear violations; 0.5, minor ordering issues; 1, logically sequenced plan.
    \end{itemize}
    \end{notebox}

    \clearpage
    \subsection{D3: Research Plan and Feasibility Justification (15 pts)}
    \begin{notebox}{D3: Research Plan and Feasibility Justification}
    This dimension evaluates the scientific rigor of research methods, clarity of the technical roadmap, visual presentation, and authenticity and consistency of prior results.

    \noindent\textbf{D3.1 Feasibility and Risk Analysis (5 pts).}
    \begin{itemize}\setlength{\itemsep}{0pt}\setlength{\parsep}{0pt}\setlength{\topsep}{0pt}
        \item \textbf{R3.1.1.} Does the feasibility analysis provide concrete evidence from preliminary experiments, prior results, or theoretical foundations? Scoring: 0, no feasibility evidence; 0.5, at least one specific number from a published team result or externally cited pilot experiment; 1, specific preliminary results directly supporting the core innovations.
        \item \textbf{R3.1.2.} Are key technical bottlenecks identified with corresponding coping strategies for each? Scoring: 0, no bottleneck analysis; 0.5, bottlenecks named but no strategies; 1, each bottleneck paired with a mitigation approach.
        \item \textbf{R3.1.3.} For the highest-risk task or core innovation, is a concrete alternative technical route or fallback measure proposed? Scoring: 0, no risk mitigation described; 0.5, risk awareness expressed but alternatives are not technically distinct; 1, at least the highest-risk task has a concrete algorithmic-level alternative that would address failure of the primary route.
        \item \textbf{R3.1.4.} Are the instruments, data resources, and software and hardware working conditions described with availability confirmed? Scoring: 0, not described; 0.5, described but availability unclear; 1, concrete resources with confirmed access.
        \item \textbf{R3.1.5.} Is method selection supported by literature or theoretical foundations rather than unsupported effectiveness claims? Scoring: 0, unsupported method choices; 0.5, weak justification; 1, strong literature and theory backing for all methods.
    \end{itemize}

    \noindent\textbf{D3.2 Technical Roadmap and Experimental Design (5 pts).}
    \begin{itemize}\setlength{\itemsep}{0pt}\setlength{\parsep}{0pt}\setlength{\topsep}{0pt}
        \item \textbf{R3.2.1.} Is there a specific technical roadmap for each sub-direction rather than a single generic overview? Scoring: 0, no per-sub-direction roadmap; 0.5, present but superficial; 1, detailed technical steps for each sub-direction.
        \item \textbf{R3.2.2.} Do the research tasks in the roadmap correspond one-to-one with all sub-directions, with no omissions? Scoring: 0, missing sub-directions; 0.5, partial coverage; 1, complete correspondence.
        \item \textbf{R3.2.3.} Is there a concrete experimental design including dataset selection, baseline methods, and validation approaches? Scoring: 0, fewer than 2 elements; 0.5, 2 elements; 1, all 3 elements with specificity.
        \item \textbf{R3.2.4.} Are quantitative metrics or operable validation standards defined for evaluating outcomes? Scoring: 0, no metrics; 0.5, generic metrics; 1, specific, measurable validation criteria.
        \item \textbf{R3.2.5.} Are all datasets and baselines cited in the experimental design traceable to specific references? Scoring: 0, unverifiable datasets and baselines; 0.5, mostly traceable; 1, all traceable to references.
    \end{itemize}

    \noindent\textbf{D3.3 Visual Technical Presentation (3 pts).}
    All items require actual embedded figures and score 0 if no figures are embedded.
    \begin{itemize}\setlength{\itemsep}{0pt}\setlength{\parsep}{0pt}\setlength{\topsep}{0pt}
        \item \textbf{R3.3.1.} Is there a Route-type figure covering the overall research framework with ordered steps and key decision nodes? Scoring: 0, no route figure; 0.5, present but unclear ordering; 1, clear sequential roadmap with milestones.
        \item \textbf{R3.3.2.} Does a Framework-type figure show all major technical modules and their information flow and dependency relationships? Scoring: 0, no framework figure; 0.5, present but incomplete; 1, comprehensive architecture with clear data flow.
        \item \textbf{R3.3.3.} Does at least one figure present the core method's specific workflow with processing steps and input and output nodes? Scoring: 0, no workflow figure; 0.5, present but vague; 1, detailed execution pipeline.
    \end{itemize}

    \noindent\textbf{D3.4 Authenticity and Consistency (2 pts).}
    \begin{itemize}\setlength{\itemsep}{0pt}\setlength{\parsep}{0pt}\setlength{\topsep}{0pt}
        \item \textbf{R3.4.1.} Are newly coined model and algorithm names defined and verifiably distinct from existing methods in cited literature? Scoring: 0, name collision or undefined; 0.5, defined but similarity unclear; 1, clearly defined and distinct.
        \item \textbf{R3.4.2.} Do prior results cited in the feasibility analysis match the research foundation section in venue, year, and performance figures? Scoring: 0, clear discrepancies; 0.5, minor inconsistencies; 1, perfect consistency.
    \end{itemize}
    \end{notebox}

    \subsection{D4: Research Foundation and Team Capability (9 pts)}
    \begin{notebox}{D4: Research Foundation and Team Capability}
    This dimension evaluates the degree of innovation, prior research accumulation, team capability, and project status authenticity.

    \noindent\textbf{D4.1 Foundation-Innovation Match (5 pts).}
    \begin{itemize}\setlength{\itemsep}{0pt}\setlength{\parsep}{0pt}\setlength{\topsep}{0pt}
        \item \textbf{R4.1.1.} Are at least 10 academic papers by the applicant directly related to the proposed topic? Scoring: 0, fewer than 5 papers; 0.5, 5 to 9 papers or at least 10 with weak relevance; 1, at least 10 papers with strong topical alignment.
        \item \textbf{R4.1.2.} Do the applicant's prior results demonstrate core technical accumulation relevant to the proposed innovations? Scoring: 0, no relevant prior work; 0.5, at least 3 team papers with direct topical overlap with the innovations, explicitly mapped in the evidence field; 1, direct, named technical continuity from at least 3 prior publications to each proposed innovation.
        \item \textbf{R4.1.3.} Are specific prior results, including datasets, models, code, and experimental conclusions, described as research foundation? Scoring: 0, no specific artifacts; 0.5, named but not detailed; 1, concrete, reusable artifacts described.
        \item \textbf{R4.1.4.} Is there evidence that the team has successfully tackled similar technical challenges in the past? Scoring: 0, no evidence; 0.5, weak evidence; 1, clear track record of relevant problem-solving.
        \item \textbf{R4.1.5.} Do the claimed innovations build naturally on the team's existing capability, or do they require a leap into unfamiliar territory? Scoring: 0, requires implausible leap; 0.5, moderate stretch; 1, natural extension of demonstrated capability.
    \end{itemize}

    \noindent\textbf{D4.2 Project Status and Authenticity (4 pts).}
    The D4.2 total is capped at twice the D7.3 score, so the full 4 points are reachable only when D7.3 reaches its maximum.
    \begin{itemize}\setlength{\itemsep}{0pt}\setlength{\parsep}{0pt}\setlength{\topsep}{0pt}
        \item \textbf{R4.2.1.} Is the status of currently funded related projects described, or explicitly stated as none? Scoring: 0, not mentioned or placeholder approval numbers; 0.5, mentioned but vague; 1, clear status with real project numbers and titles.
        \item \textbf{R4.2.2.} Are previously completed NSFC projects described, or explicitly stated as none? Scoring: 0, not mentioned; 0.5, mentioned but vague; 1, clear completion status with outcomes.
        \item \textbf{R4.2.3.} Are all publication venues in the applicant's prior work list real and verifiable? Scoring: 0, fictitious venues detected; 0.5, mostly verifiable; 1, all venues verifiable.
        \item \textbf{R4.2.4.} Are all publication years logically valid, with in-press status explicitly stated? Scoring: 0, invalid years present; 0.5, minor issues; 1, all years valid.
    \end{itemize}
    \end{notebox}

    \subsection{D5: Writing Standards and Logic Presentation (4 pts)}
    \begin{notebox}{D5: Writing Standards and Logic Presentation}
    This dimension evaluates format compliance, language rigor, and overall logical coherence.
    \begin{itemize}\setlength{\itemsep}{0pt}\setlength{\parsep}{0pt}\setlength{\topsep}{0pt}
        \item \textbf{R5.1.} Is the total page count at most 45 pages? Scoring: 0, more than 45 pages; 1, at most 45 pages.
        \item \textbf{R5.2.} Does the five-element logic chain from problem to objectives to content to plan to innovations form a traceable thread throughout the document? Scoring: 0, disconnected sections; 0.5, weak connections; 1, clear, traceable logic chain.
        \item \textbf{R5.3.} Does the reference format comply with GB/T 7714 with author names present? Scoring: 0, non-compliant or authors missing; 0.5, minor format issues; 1, fully compliant.
        \item \textbf{R5.4.} Is formal academic language used with all abbreviations expanded on first use? Scoring: 0, colloquial or unexpanded abbreviations; 0.5, mostly formal; 1, consistently formal and precise.
    \end{itemize}
    \end{notebox}

    \subsection{D6: Figure Quality (5 pts)}
    \begin{notebox}{D6: Figure Quality}
    This dimension evaluates the quality of embedded figures and figure-text coordination. All items require actual embedded figures, and all items score 0 if no figures are found.
    \begin{itemize}\setlength{\itemsep}{0pt}\setlength{\parsep}{0pt}\setlength{\topsep}{0pt}
        \item \textbf{R6.1.} Are all embedded figures of readable resolution with clearly legible text and graphical elements? Scoring: 0, unreadable or no figures; 0.5, readable but marginal quality; 1, high-quality, crisp figures.
        \item \textbf{R6.2.} Does at least one figure serve a Motivation function, visually presenting existing method limitations or gaps cross-validated with the rationale? Scoring: 0, no motivation figure; 0.5, figure exists but the contrast is unclear or implicit; 1, clear visual argument showing the gap or limitation.
        \item \textbf{R6.3.} Does at least one figure serve a Framework function, showing the architecture with all major modules and information flow? Scoring: 0, no framework figure; 0.5, present but incomplete; 1, comprehensive architecture diagram.
        \item \textbf{R6.4.} Does at least one figure serve a Route function, presenting ordered research steps consistent with the plan? Scoring: 0, no route figure; 0.5, present but unclear ordering; 1, clear sequential roadmap.
        \item \textbf{R6.5.} For figures explicitly referenced in text, does the visual content match the textual description with no contradictions? Scoring: 0, contradictions present; 0.5, weak match; 1, perfect figure-text consistency.
    \end{itemize}
    \end{notebox}

    \subsection{D7: Citation Credibility and Fact-Checking (10 pts)}
    \begin{notebox}{D7: Citation Credibility and Fact-Checking}
    This dimension evaluates citation credibility through active web search verification. Every item requires a completed and logged web search, and a score of 1 is never awarded without positive confirmation.

    \noindent\textbf{D7.1 Core Citation Existence Verification (3 pts).}
    \begin{itemize}\setlength{\itemsep}{0pt}\setlength{\parsep}{0pt}\setlength{\topsep}{0pt}
        \item \textbf{R7.1.1.} From the rationale section, randomly select 5 key references accompanying quantitative claims and verify each exists on arXiv, Semantic Scholar, or DBLP. Scoring: 0, at most 2 of 5 papers confirmed real; 0.5, exactly 3 confirmed; 1, at least 4 confirmed with identifiers found.
        \item \textbf{R7.1.2.} For all references dated 2024 to 2025, verify they were actually published in the stated venue. Scoring: 0, at least one paper cannot be found in any public database; 0.5, all found but with nontrivial mismatches; 1, all 2024 to 2025 citations fully verified.
        \item \textbf{R7.1.3.} Verify all conference and journal names in the reference list are real academic venues. Scoring: 0, at least one venue cannot be confirmed; 0.5, all real but some mischaracterized; 1, all venues confirmed and correctly categorized.
    \end{itemize}

    \noindent\textbf{D7.2 Quantitative Claim Verification (3 pts).}
    \begin{itemize}\setlength{\itemsep}{0pt}\setlength{\parsep}{0pt}\setlength{\topsep}{0pt}
        \item \textbf{R7.2.1.} Select the 3 most specific quantitative claims in the proposal and verify the exact numbers appear in the source papers. Scoring: 0, none verified; 0.5, source papers found but numbers differ by more than 3 percent absolute; 1, at least 2 of 3 claims precisely confirmed.
        \item \textbf{R7.2.2.} For all benchmark datasets named in the experimental design, verify they are real datasets with task definitions matching the proposal's description. Scoring: 0, at least one cannot be found; 0.5, datasets exist but task descriptions differ; 1, all datasets verified and consistent.
        \item \textbf{R7.2.3.} For all market or industry statistics, verify traceability to a named authoritative source. Scoring: 0, cited with no reference and numbers not found in any credible source; 0.5, named source category but no reference number, or similar data with moderate differences; 1, all statistics have a named, locatable source; score 1 if no market statistics appear.
    \end{itemize}

    \noindent\textbf{D7.3 Applicant Prior Work Verification (2 pts).}
    \begin{itemize}\setlength{\itemsep}{0pt}\setlength{\parsep}{0pt}\setlength{\topsep}{0pt}
        \item \textbf{R7.3.1.} Search the applicant's full name and institution on DBLP, then verify 3 recent publications match in author, title, year, and venue. Scoring: 0, applicant not found or at least 2 of 3 papers not found; 0.5, profile found but details mismatch; 1, profile found and all 3 papers fully confirmed.
        \item \textbf{R7.3.2.} For performance numbers the applicant claims for their own prior work, verify they appear in the original paper. Scoring: 0, claimed number absent; 0.5, number found but in a different experimental context; 1, number exactly matches the abstract or primary results table.
    \end{itemize}

    \noindent\textbf{D7.4 Technical Claims versus Field State-of-the-Art (2 pts).}
    \begin{itemize}\setlength{\itemsep}{0pt}\setlength{\parsep}{0pt}\setlength{\topsep}{0pt}
        \item \textbf{R7.4.1.} Verify that the stated best baseline performance is consistent with publicly known state of the art. Scoring: 0, more than 5 percent absolute below the actual state of the art; 0.5, plausible but stale references; 1, consistent with 2024 to 2025 results.
        \item \textbf{R7.4.2.} For any first-ever or domestic-first claims, confirm no highly similar published work exists prior to this proposal. Scoring: 0, highly similar work found and neither cited nor distinguished; 0.5, related work found but the approach has clear differentiation; 1, no close prior work found.
    \end{itemize}
    \end{notebox}

    \clearpage
    \subsection{D8: Internal Consistency Deep Check (6 pts)}
    \begin{notebox}{D8: Internal Consistency Deep Check}
    This dimension evaluates internal consistency across sections. Every item requires explicit cross-section comparison, quoting both text locations in the evidence field.
    \begin{itemize}\setlength{\itemsep}{0pt}\setlength{\parsep}{0pt}\setlength{\topsep}{0pt}
        \item \textbf{R8.1.} Does every quantitative metric in the research objectives have a corresponding verification arrangement in the annual plan? Scoring: 0, at least one target has no corresponding verification; 0.5, arrangements exist but the metric does not match; 1, every target has a traceable annual verification node.
        \item \textbf{R8.2.} Do self-citations in the feasibility analysis match the research foundation section in venue, year, and performance figures? Scoring: 0, same paper described differently; 0.5, numbers consistent but minor descriptive differences; 1, all cross-section self-citations perfectly consistent.
        \item \textbf{R8.3.} Does each claimed innovation have a dedicated verification experiment in the experimental design? Scoring: 0, at least one innovation lacks one; 0.5, experiment exists but cannot isolate the innovation's contribution; 1, each innovation has a dedicated ablation or controlled comparison.
        \item \textbf{R8.4.} Are all key technical terms defined at first use and used consistently throughout the document? Scoring: 0, a core term is undefined or its meaning shifts; 0.5, 1 to 2 abbreviations lack expansion but meaning is inferable; 1, all terms defined and consistently applied.
        \item \textbf{R8.5.} When the same baseline is stated in both the rationale and the research objectives, are the numbers identical? Scoring: 0, different numbers; 0.5, numbers match but conditions described inconsistently; 1, identical in value and conditions.
        \item \textbf{R8.6.} Can every research sub-direction be traced to a research gap, a technical route, and a verifiable expected output? Scoring: 0, any sub-direction is missing at least one link; 0.5, all have 2 of 3 links; 1, all three links confirmed with explicit text evidence.
    \end{itemize}
    \end{notebox}

    In addition to the 78 items above, the rubric defines pre-check rules: if no figures are embedded, items R1.4.1, R3.3.1, R3.3.2, R3.3.3, and all of R6 are scored 0; if the total page count exceeds 45, R5.1 is scored 0; and if the ongoing-projects table contains placeholder text, R4.2.1 is scored 0. Anti-hallucination rules require graders to verify figure presence, trace quantitative claims, require keyword tracing for logic chains, and compare cross-section values explicitly rather than relying on overall impression.

\section{Human Evaluation Protocol}
\label{app:human_eval}

\paragraph{Participants and setup.}
We recruited several NSFC applicants and grant review experts who have direct experience writing or evaluating NSFC proposals. All participants are colleagues of the authors' collaborators; they were invited by the research team and participated voluntarily with no compensation or coercion. Informed consent was obtained from each participant prior to the evaluation. Participants were shown a set of NSFC proposals generated on the same research topic in randomized, anonymized order. They were instructed not to search online and to evaluate each proposal independently before providing an intra-topic ranking.

\paragraph{Simplified rubric.}
To reduce reviewer burden while preserving dimensional alignment with the LLM rubric, we condensed the 78-item binary checklist into six holistic dimensions scored on a 1--5 integer scale (1 = very poor, 3 = acceptable, 5 = excellent). The mapping from the LLM rubric dimensions to the human rubric dimensions is: D1 (Scientific Significance) $\leftarrow$ LLM D1; D2 (Innovation) $\leftarrow$ LLM D2; D3 (Research Content and Objectives) $\leftarrow$ LLM D2+D3; D4 (Plan Feasibility) $\leftarrow$ LLM D3+D4; D5 (Research Foundation) $\leftarrow$ LLM D4+D5; D6 (Visual Quality) $\leftarrow$ LLM D6.

Two LLM dimensions are omitted due to prohibitive annotation cost. \textbf{D7 (Citation Credibility)} requires an active, logged web search for every item---verifying each reference on arXiv, Semantic Scholar, or DBLP, tracing quantitative claims back to source papers, and checking the applicant's prior work---a workload that would multiply reviewer time by several-fold and is infeasible in a blind offline setting. \textbf{D8 (Internal Consistency)} demands exhaustive cross-section number comparisons, which alone can take longer than scoring the remaining six dimensions combined; more fundamentally, several D8 items depend directly on the citation-level ground truth that D7 establishes---without first completing D7, those D8 sub-items cannot be evaluated at all, making D8 logically downstream of D7. Human scores are therefore reported over a 30-point total and compared with LLM dimension-wise scoring rates.

\paragraph{Scoring rubric instructions.}
Reviewers received the following instructions for each dimension. All dimensions use a 1--5 integer scale: 1 = very poor or absent; 2 = weak, content is formulaic or superficial; 3 = acceptable, meets basic requirements with minor gaps; 4 = good, concrete but lacking depth; 5 = excellent, specific and in-depth with strong coupling to the full proposal. Reviewers were encouraged to use the full range and to avoid clustering scores at 3. The overall score need not equal the average of dimension scores; a critical weakness in any dimension should pull the overall score down.

\begin{itemize}\setlength{\itemsep}{2pt}
  \item \textbf{D1 Scientific Significance.} Does the proposal identify a clear research gap? Is the problem important? A score of 5 requires a specific bottleneck with supporting data, policy references, or scenarios and a critical characterisation of why existing work falls short. A score of 1 indicates generic motivation (``with the development of X\ldots'') with no clear rationale for why the work is worth doing.
  \item \textbf{D2 Innovation.} Can the novelty be identified at a glance relative to existing work? A score of 5 requires an explicit comparison to the closest existing work and a layered set of innovations. A score of 1 indicates no clear innovation or the relabelling of standard practice as novel. Claiming ``we will propose X'' without comparison to existing work caps the score at 2.
  \item \textbf{D3 Research Content and Objectives.} Are objectives measurable and verifiable? Is sub-topic decomposition logical? A score of 5 requires quantifiable objectives with explicit metrics, logically progressive and non-overlapping sub-topics, and scientific questions that explain why existing methods fail. A score of 1 indicates vague objectives, disorganised sub-topics, or scientific questions that merely restate the research content.
  \item \textbf{D4 Plan Feasibility.} Are technical methods described with sufficient detail? Are experimental means complete? A score of 5 requires method-level detail (model, algorithm, or formula), complete experimental design (data, baselines, metrics), and identified risks with mitigation strategies. A score of 1 indicates only high-level slogans with no technical specifics.
  \item \textbf{D5 Research Foundation.} How well does prior work align with the proposed project? Are working conditions specific? A score of 5 requires directly relevant preliminary results and verifiable working conditions matched to the project. A score of 1 indicates a mismatch between prior work and the proposal, or conditions filled with placeholder text.
  \item \textbf{D6 Visual Quality.} Is the layout neat? Are figures clear and well-integrated with the text? A score of 5 indicates polished layout and figures that are crisp, captioned, and tightly coupled to the surrounding text. A score of 1 indicates disorganised layout or low-quality figures that are blurry, uncaptioned, or irrelevant.
\end{itemize}

\section{System Comparison}
\label{app:system_comparison}
            
\begin{table}[h!]
\centering
\small
\resizebox{\textwidth}{!}{
\begin{tabular}{lcccccc}
\toprule
\textbf{System} & \textbf{Target Output} & \textbf{Planning Strategy} & \textbf{Template Adaptation} & \textbf{Literature \& Analysis} & \textbf{Author Grounding} & \textbf{Human-in-the-Loop} \\
\midrule
ResearchAgent~\cite{baek2024researchagent} & Research Ideas & Iterative refinement & Fixed JSON schema & KG \& entity-based search & N/A (Idea focus) & None (autonomous) \\
AutoSurvey~\cite{wang2024autosurvey}       & Survey Papers  & Hierarchical outline & Static \LaTeX{} style & Topic-clustered RAG & N/A (Public literature) & Autonomous \\
CycleResearcher~\cite{weng2025cycleresearcher} & Full Manuscripts & Section-sequential & Fixed conference style & ArXiv keyword search & N/A (Standard drafting) & Autonomous review \\
ARIS~\cite{yang2026aris}                   & Full Manuscripts & Modular skill graph & Predefined workflow & Reusable search skills & N/A (Experiment focus) & Multi-agent loop \\
\midrule
\textbf{FundingClaw (Ours)} & \textbf{Grant Proposals} & \textbf{Budgeted Blueprint} & \textbf{Decoupled Parser} & \textbf{Prefetch + Causal Memo} & \textbf{Multi-DB + PDF Parsing} & \textbf{Stage Checkpoints} \\
\bottomrule
\end{tabular}
}
\caption{\textbf{Comparison of representative scientific writing and ideation agent systems across key architectural dimensions.}
Unlike existing systems tailored for open-ended ideation or conventional paper drafting, \textbf{FundingClaw} explicitly addresses the regulatory heterogeneity (multi-agency templates), strict length/budget constraints, and zero-hallucination applicant-record verification unique to grant proposal writing.}
\label{tab:system_comparison}
\end{table}

\end{document}